\documentclass[conference]{IEEEtran}
\IEEEoverridecommandlockouts  

\usepackage{multirow}
\usepackage{amsmath,amssymb,amsthm}
\usepackage{algorithm}
\usepackage{algorithmic}
\usepackage{listings}
\usepackage{xcolor}
\usepackage{graphicx}
\usepackage{booktabs}
\usepackage{hyperref}
\usepackage{url}

\lstdefinelanguage{json}{
    basicstyle=\ttfamily\small,
    stringstyle=\color{blue},
    numberstyle=\scriptsize,
    stepnumber=1,
    numbersep=8pt,
    showstringspaces=false,
    breaklines=true,
    literate=
     *{0}{{{\color{purple}0}}}{1}
      {1}{{{\color{purple}1}}}{1}
      {2}{{{\color{purple}2}}}{1}
      {3}{{{\color{purple}3}}}{1}
      {4}{{{\color{purple}4}}}{1}
      {5}{{{\color{purple}5}}}{1}
      {6}{{{\color{purple}6}}}{1}
      {7}{{{\color{purple}7}}}{1}
      {8}{{{\color{purple}8}}}{1}
      {9}{{{\color{purple}9}}}{1}
      {:}{{{\color{black}{:}}}}{1}
      {,}{{{\color{black}{,}}}}{1}
      {\{}{{{\color{black}{\{}}}}{1}
      {\}}{{{\color{black}{\}}}}}{1}
      {[}{{{\color{black}{[}}}}{1}
      {]}{{{\color{black}{]}}}}{1},
}

\newcommand{\bibstylename}{IEEEtran}

\newtheorem{definition}{Definition}

\makeatletter
\renewcommand\paragraph{\@startsection{paragraph}{4}{1em}%
  {1ex plus .2ex minus .1ex}%
  {-0.5em}%
  {\normalfont\normalsize\itshape}}
\makeatother

\begin{document}

\title{HYVE: Hybrid Views for LLM
Context \\Engineering over Machine Data}

\author{
\IEEEauthorblockN{Jian Tan$^*$, Fan Bu$^*$\thanks{$^*$Equal contribution.}, Yuqing Gao, Dev Khanolkar, Jason Mackay, Boris Sobolev, Lei Jin, Li Zhang}
\IEEEauthorblockA{Cisco Systems, Inc., San Jose, California, USA\\
\{jianta,fabu,yuqingga,jasmacka,dkhanolk,bsobolev,leiji2,liz4\}@cisco.com}
}

\maketitle

\begin{abstract}
 Machine data is central to observability and diagnosis in modern computing systems, appearing in logs, metrics, telemetry traces, and configuration
  snapshots. When provided to large language models (LLMs), this data typically arrives as a mixture of natural language and structured payloads such as
  JSON or Python/AST literals. Yet LLMs remain brittle on such inputs, particularly when they are long, deeply nested, and dominated by repetitive
  structure.

  We present HYVE (HYbrid ViEw), a framework for LLM context engineering for inputs containing large machine-data payloads,
  inspired by database management principles. HYVE
  surrounds model invocation with coordinated preprocessing and postprocessing, centered on a request-scoped datastore augmented with schema information.
  During preprocessing, HYVE detects repetitive structure in raw inputs, materializes it in the datastore, transforms it into hybrid columnar and row-
  oriented views, and selectively exposes only the most relevant representation to the LLM. During postprocessing, HYVE either returns the model output
  directly, queries the datastore to recover omitted information, or performs a bounded additional LLM call for SQL-augmented semantic synthesis.

  We evaluate HYVE on diverse real-world workloads spanning knowledge QA, chart generation, anomaly detection, and multi-step network troubleshooting.
  Across these benchmarks, HYVE reduces token usage by 50--90\% while maintaining or improving output quality. On structured generation tasks, it improves chart-generation accuracy by up to 132\% and reduces latency by up to 83\%. Overall, HYVE offers a practical approximation to an effectively unbounded
  context window for prompts dominated by large machine-data payloads.
\end{abstract}

\section{Introduction}
\label{sec:intro}

Machine data, such as logs, metrics, telemetry traces, and configuration snapshots, is ubiquitous in modern computing systems and underpins
  observability. When fed into LLM prompts or carried across multi-turn interactions, it often appears as an interleaving of natural language
  and large structured payloads.
LLMs continue to struggle with such inputs for three recurring reasons:
  \begin{itemize}
  \item \textbf{Token explosion from verbosity:} Nested keys and repeated schema consume the context window, fragmenting relevant evidence and
  crowding out useful data.
  \item \textbf{Context rot:} The model misses the ``needle'' buried in large payloads and drifts from the instruction, losing track of task-relevant signals.
  \item \textbf{Weaknesses in numeric and categorical sequence reasoning:} Long sequences obscure patterns such as anomalies, trends, and
  entity relationships that are critical for data analytics.
  \end{itemize}

\begin{figure}[hbt]
\centering
\includegraphics[width=0.49\textwidth]{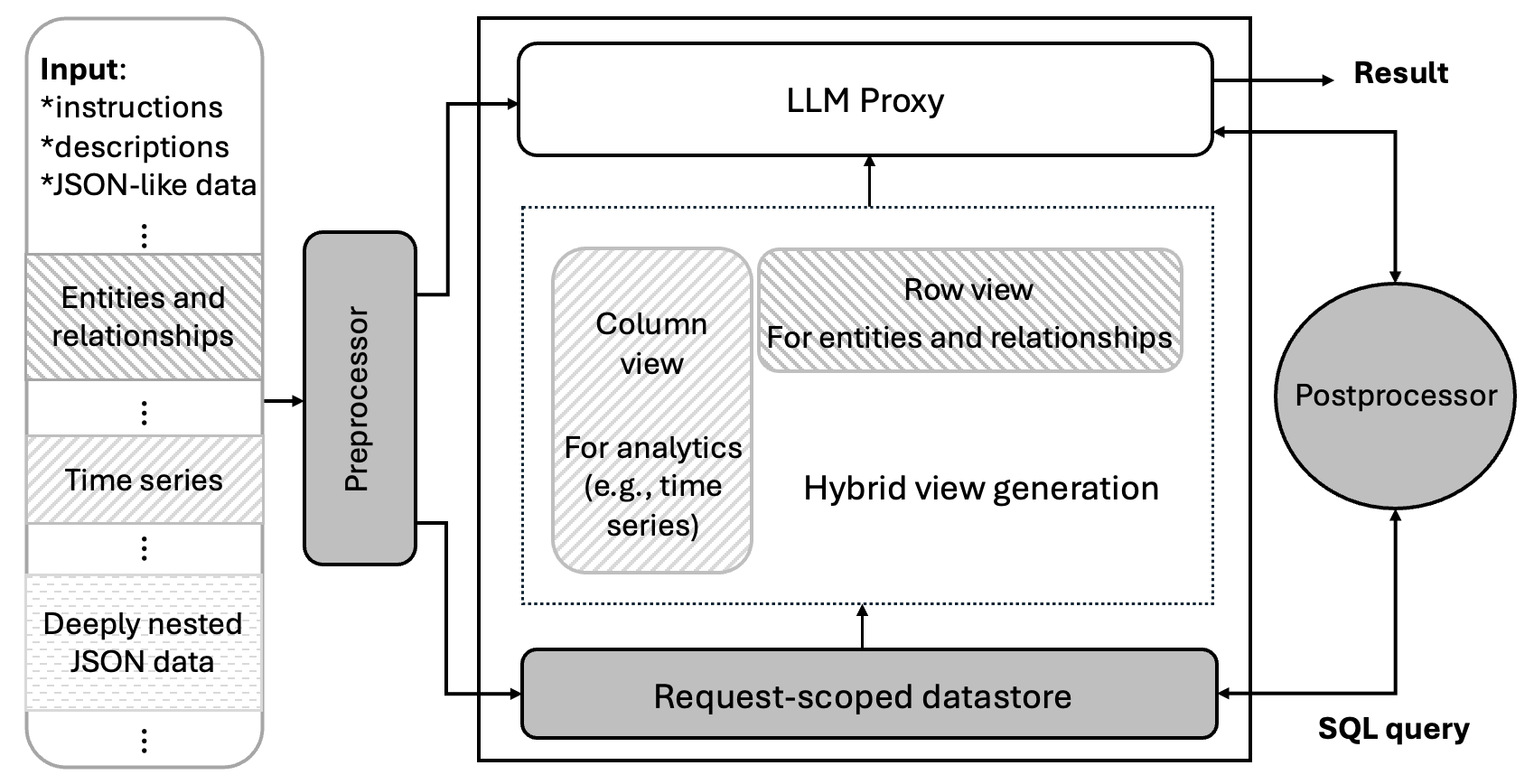}
\caption{HYVE architecture. The preprocessor parses input strings with embedded structured segments into hybrid tabular views; the
  request-scoped datastore retains the full data; and the postprocessor inspects LLM outputs and, when needed, queries the datastore to enrich
  the response.}
\label{fig:architecture}
\end{figure}

Machine data often becomes long because system code naturally emits repeated structures: loops generate arrays, repeated keys, and nested records in the observable representation. The bottleneck is therefore not raw length alone. Effective use of such data requires \emph{structural transformation} and \emph{signal enhancement}, so that the same information is presented in forms better aligned with the reasoning strengths of LLMs~\cite{aceblog}\footnote{HYVE was initially developed as DNM-ACE (Analytics Context Engineering for the Cisco Deep Network Model) and later renamed DNM-HYVE. Section~\ref{sec:dnm} discusses its synergy with the DNM model.}. Empirically, we observe that, in long-context settings, LLMs still make errors when correlating multiple arrays through their indices, whereas a row-oriented view makes this easier by presenting related values side by side. Conversely, LLMs often struggle to extract values from the same field across a list of dictionaries, such as when generating a line chart, whereas a column-oriented view places those values consecutively in order.

\subsection{Motivating examples}
\begin{figure}[t]
\centering
\includegraphics[width=0.98\columnwidth]{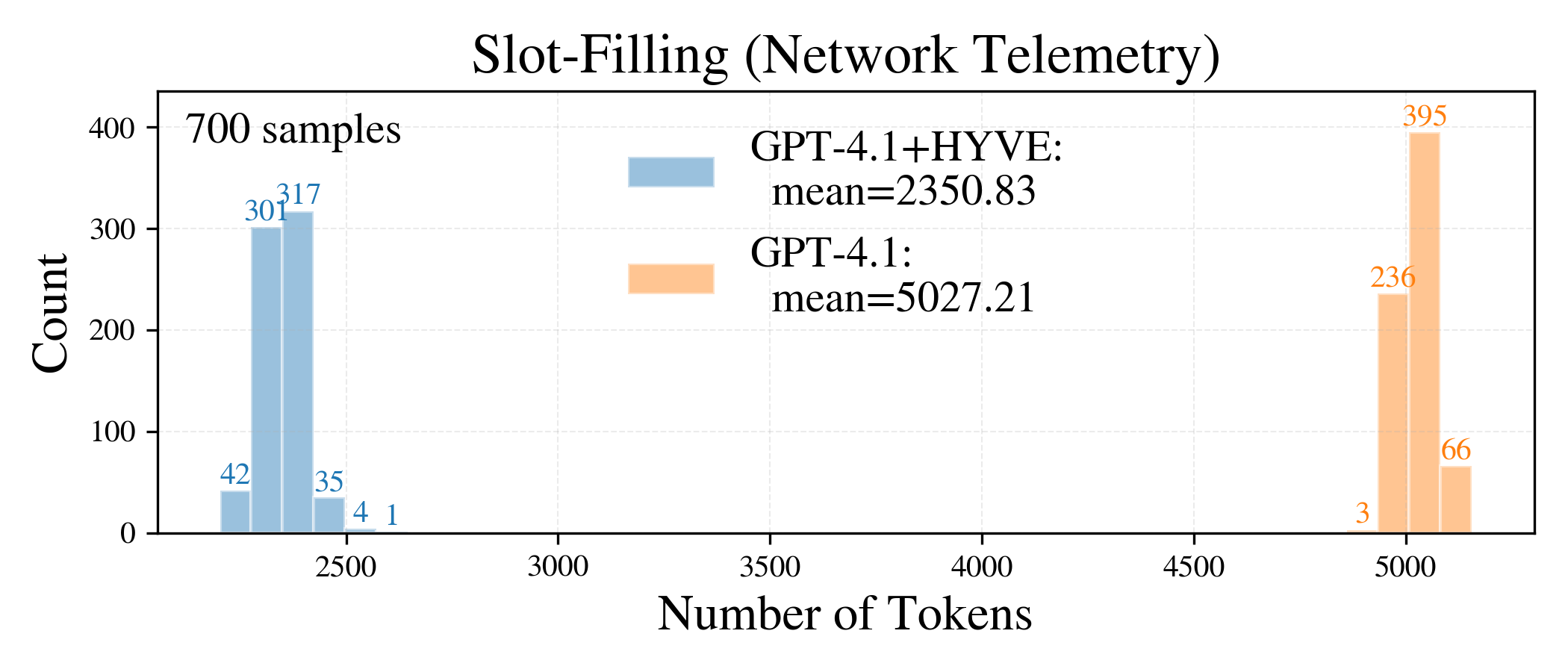}\par
\vspace{2mm}
\includegraphics[width=0.98\columnwidth]{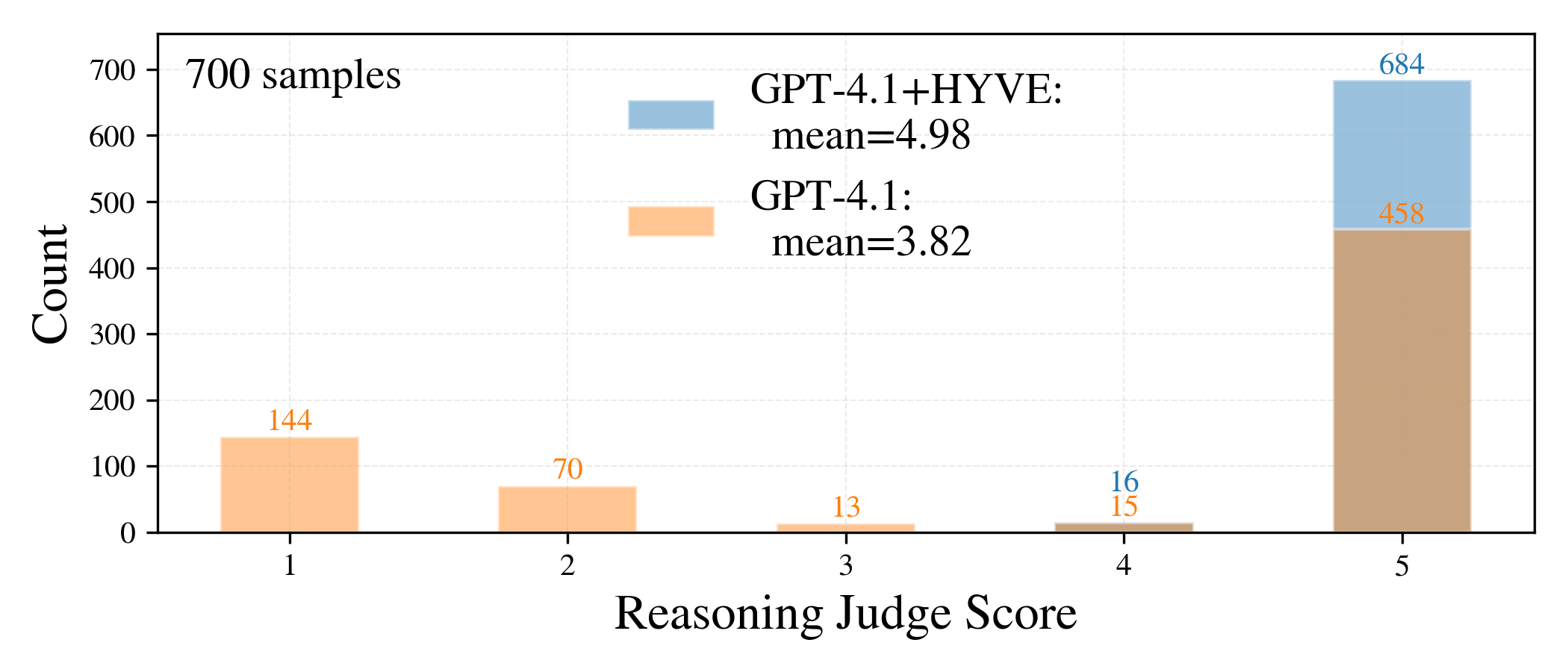}\par
\caption{Slot-filling tasks over 700 network telemetry samples. Top: baseline GPT-4.1 consumes ${\sim}$5000 tokens per request; HYVE reduces usage by 53\%. Bottom: baseline GPT-4.1 fails on 30.6\% of samples (scores $\le$2); HYVE achieves almost perfect scores on all samples.}
\label{fig:motivating-canvas}
\end{figure}

\begin{figure}[t]
\centering
\includegraphics[width=0.98\columnwidth]{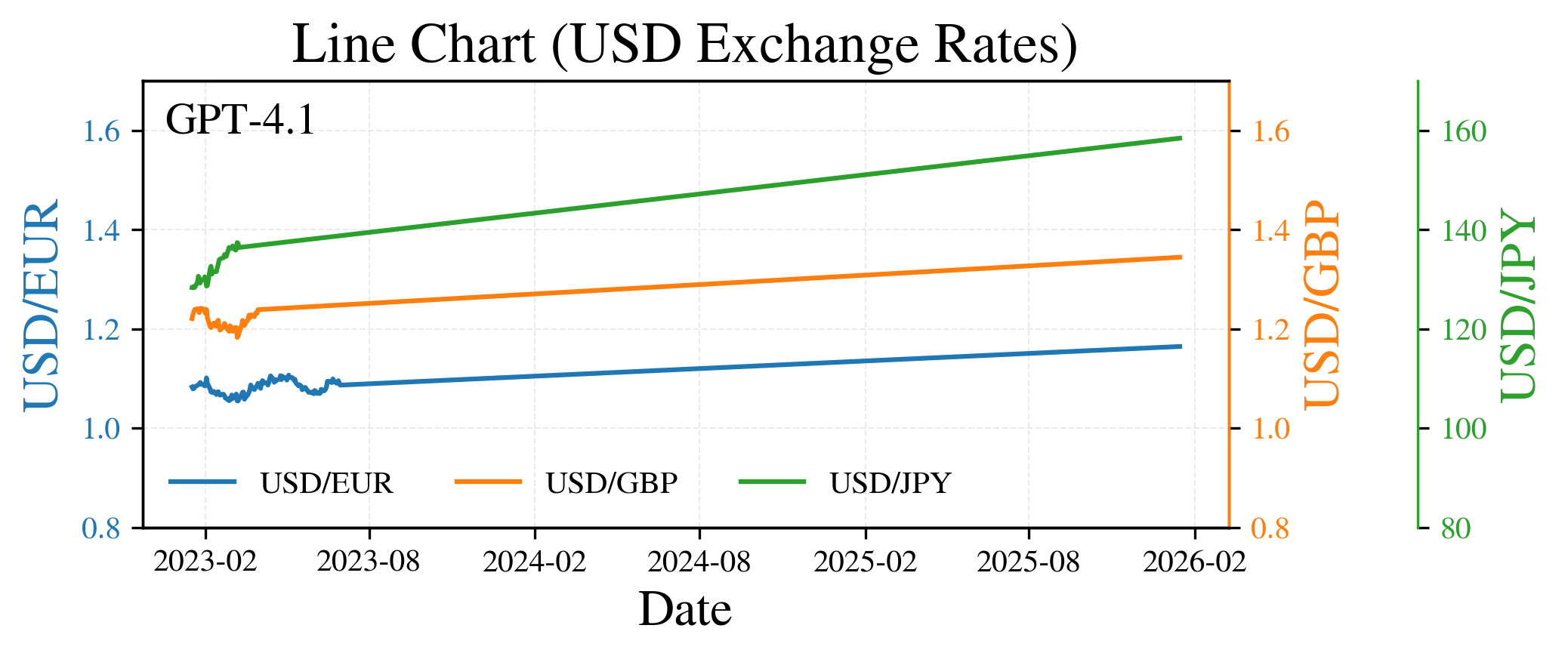}\par
\vspace{2mm}
\includegraphics[width=0.98\columnwidth]{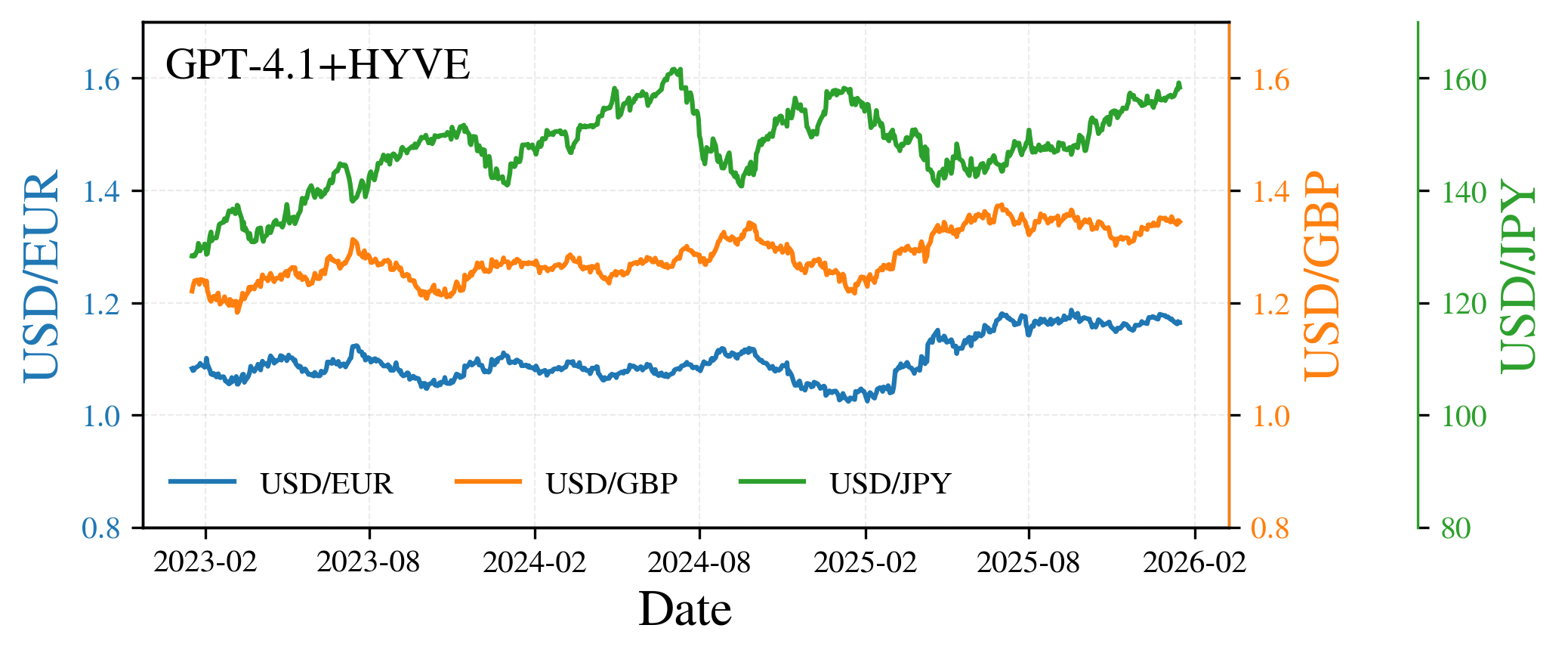}\par
\caption{Line chart over three years of USD exchange-rate data (778 points per series). Baseline GPT-4.1 truncates output arrays to 40--120 points,
  collapsing detailed time series into sparse straight-line segments. HYVE recovers all 778 points through data backfilling.}
\label{fig:motivating-forex}
\end{figure}

We illustrate the problem with two real examples. Figure~\ref{fig:motivating-canvas} shows a slot-filling task, i.e., semantic querying over network telemetry data subject to specified requirements. Baseline GPT-4.1 consumes more than 5,000 tokens per request yet fails on 30.6\% of samples (scores $\le$2), showing that naively embedding raw machine data in prompts is both inefficient and unreliable. In contrast, HYVE reduces token usage by 53\% while achieving near-perfect quality (mean score 4.98) on all 700 test cases. Figure~\ref{fig:motivating-forex} shows a second example: when asked to generate a line chart from 778 exchange-rate points, baseline GPT-4.1 truncates output arrays to 40--120 points, causing most of the data to disappear and collapsing detailed time series into sparse straight-line segments. With HYVE, all 778 points are recovered, producing a faithful chart. HYVE has been deployed in Cisco AI products and is effective across a broad range of machine-data analytics tasks.

\subsection{Guiding principles}

HYVE is guided by database-inspired principles for analytics-oriented context engineering. It treats prompt construction not as ad hoc compression, but as a disciplined process of structure discovery, information preservation, and queryable representation.
  \begin{itemize}
  \item \textbf{Guarantee data fidelity:} Context reduction must not lose information. Every part of the raw input is either preserved verbatim in the visible LLM context or retained in a request-scoped datastore with explicit schema information, so that omitted content remains fully recoverable.
  \item \textbf{Avoid semantic compression:} Context reduction does not rely on LLM-based summarization, 
  which may be ineffective for machine data. Instead, HYVE detects repetitive structures, organizes them into tables with precise schema, and exposes only a subset of entries selected by statistical ranking,
  while explicitly indicating that additional data remain available in the datastore. When hidden data are needed, the LLM can generate SQL over the provided schema, after which the postprocessor either backfills the result into a template or performs one additional synthesis step.
  \item \textbf{Limit LLM calls:} HYVE is not an open-ended iterative agent. Instead, it operates as a service that preserves the same interface as a standard LLM endpoint. Beyond the
  primary LLM call, the system makes at most one additional LLM call by default; this can be configured to a small fixed number for SQL refinement, and
  is only needed when semantic synthesis over the retrieved evidence is required.
  \end{itemize}

 Accurately generating SQL in a general open-ended environment remains challenging for LLMs. Our setting, however, is controlled: the
  preprocessor and postprocessor are tightly coupled within the scope of a single request. HYVE exploits this local request context to guide the
  LLM toward generating high-quality SQL queries tailored to each request. Because this scope is intentionally confined to a single
  request, HYVE provides only short-term memory.  Section~\ref{sec:limitations} further discusses the distinction between short- and long-term memory.

 Unlike long-horizon agent memory in systems such as OpenClaw~\cite{openclaw_agent}, Claude Code~\cite{claude_code}, Codex~\cite{codex_cli}, Gemini
  CLI~\cite{gemini_cli}, OpenCode~\cite{opencode_agent}, and Pi~\cite{pi_coding_agent}, which often compress prior interaction history through
  summarization, HYVE uses exact request-scoped state for query-based recovery, making it a complementary approach.

\subsection{High-level design}
As illustrated in Figure~\ref{fig:architecture}, HYVE wraps a primary LLM call with a preprocessor and a postprocessor. By default, the postprocessor may trigger at most one additional LLM call.

The \emph{preprocessor} performs three tasks: (1) parse the raw input and detect repetitive structures in embedded JSON/AST literals; (2) build an in-memory datastore that retains the full data together with inferred schema; and (3) transform the data into hybrid column and row views, exposing only a selected subset of table entries to the LLM. The \emph{postprocessor} then operates in one of three modes: (1) return the LLM output directly when the visible context is sufficient; (2) query the datastore and backfill the retrieved data into an LLM-generated template via schema-aware mappings; or (3) query the datastore, append the results to the context, and make one additional LLM call to synthesize the final answer.
The result is a compact prompt that still preserves task-critical information: although some table entries are hidden from the visible context, the schema, selected entries, surrounding instructions, and non-tabular key-value pairs remain available to the model.

\subsection{``Everything Is a File,'' and Some Are Databases}
Although HYVE is not itself an agent, we have built a network troubleshooting agent on top of it~\cite{aceblog}. That system combines a virtual file
  system, which maps observability endpoints to files and transparently intercepts Bash tools, with database-style management of structured
  data. The file-system abstraction provides intuitive and stateful high-level organization, while the datastore provides precise control
  over fine-grained structured entries.  This paper focuses exclusively on HYVE.

Anthropic~\cite{claude} and Vercel~\cite{vercel} popularized the idea that ``bash is all you need'' and, more broadly, a CLI-native approach to agentic
  workflows built on the file system and composable shell tools. In machine-data-heavy settings, however, database principles become equally important.
  Shell tools such as \texttt{grep} and \texttt{awk} are useful for pattern matching, but they are poorly suited to structured search involving
  filtering, aggregation, and joins over multiple nested machine-data segments scattered across the input. HYVE therefore stores full-fidelity copies in
  a \emph{datastore} that supports formal query processing, while presenting hybrid views better aligned with LLM reasoning.

\section{Main design}
\label{sec:problem}

We formalize analytics-oriented context engineering for LLM inputs containing embedded semi-structured segments, using hybrid views and an on-demand delayed query process implemented through coupled preprocessors and postprocessors around an LLM call.

\subsection{Preprocessor design}
The preprocessor uses a robust parser to detect repetitive structures together with their schemas, transform them into column and row views, and selectively expose only a subset of table entries to the LLM. The full dataset remains available in a queryable datastore.
\paragraph{Raw input}
The input is a single opaque string $s$\footnote{At the API level, inputs are often represented as structured objects, for example as JSON messages conforming to the interaction schema of the target LLM service. Before inference, the provider serializes this structure into a single token stream, potentially inserting vendor-specific control or delimiter tokens. As this serialization detail does not affect the design of HYVE, we omit it for simplicity.} that may contain an arbitrary mixture of natural language and machine data, such as job descriptions, instructions, command outputs, log fragments, and
  configuration snippets. No explicit structural boundary markers are guaranteed: JSON or AST literals may appear inline, span multiple lines, be
  partially malformed, or be string-encoded within other JSON fields.

\paragraph{Parsing and structure detection}
A resilient parser decomposes $s$ into an interleaved sequence of free-form text segments and structured objects:
\[
\mathsf{Parse}(s) = (t_1, J_1, t_2, J_2, \ldots, t_m, J_m, t_{m+1}),
\]
 where each $t_i$ is a text segment and each $J_i$ is a JSON or Python literal object. The parser handles deeply nested objects, heterogeneous arrays, partially malformed JSON, string-encoded nested blobs, and Python AST literals (Section~\ref{sec:system}). We write $\mathbf{J} = (J_1, \ldots, J_m)$ for the extracted objects and $\mathbf{t} = (t_1, \ldots, t_{m+1})$ for the text segments.

Using JSON’s key-value structure together with JSONPath~\cite{jsonpath}, 
a standard path notation for addressing values inside nested JSON objects, we identify \emph{repetitive patterns}, 
such as collections of dictionaries or lists that share a common JSONPath prefix and 
a common inferred schema. For example, in a JSON object whose field \texttt{items} stores a list of records, 
the path \texttt{\$.items[0].name} refers to the \texttt{name}
  field of the first record, starting from the root node \texttt{\$} and traversing through the \texttt{items} field.
These patterns can be organized into tables with well-defined column names and 
value types; they form the basis of both the hybrid view and the queryable datastore. We use JSONPath notation as a compact way to refer to values inside nested JSON objects, and write $\mathsf{Val}(J,p)$ for the value at path $p$ when it exists.

The key requirement is that entries grouped into the same table or column family must be structurally compatible. We therefore formalize the notion of \emph{schema consistency}, which determines when multiple JSON values may be safely aligned and merged:

  \begin{definition}[Schema Consistency]
  \label{def:consistent}
  Let $L = [e_1,\ldots,e_n]$ be a list of structured JSON values. We say that $L$ is
  \emph{schema-consistent} if all elements of $L$ share the same recursive schema
  signature, i.e., they have the same field structure and compatible leaf types at
  corresponding positions. In particular, for object-valued elements, this requires
  that they have the same key set and that corresponding child fields are themselves
  schema-consistent recursively. Otherwise, $L$ is \emph{schema-inconsistent}.
  \end{definition}

\paragraph{Representation operator}
Let $\mathsf{Tok}(\cdot)$ denote the token-count function under the target tokenizer.
The system transforms the raw input $s$ into a prompt string $\pi$ that preserves sufficient information for the LLM while substantially reducing token count. We denote this preprocessing step by the \emph{representation} operator
\[
\mathsf{Rep}: s \mapsto \pi,
\]
which operates on $s$ via $\mathsf{Parse}(s)$.
It is governed by subset-selection parameters, such as the number of list items retained per array and the number of top-ranked records surfaced in the row view. In our implementation these are fixed thresholds rather than an explicit token budget, yet the resulting $\mathsf{Tok}(\pi)$ is consistently 50--90\% smaller than $\mathsf{Tok}(s)$ across all evaluated workloads (Section~\ref{sec:eval}).

\begin{figure}[hbt]
\centering
\includegraphics[width=0.34\textwidth]{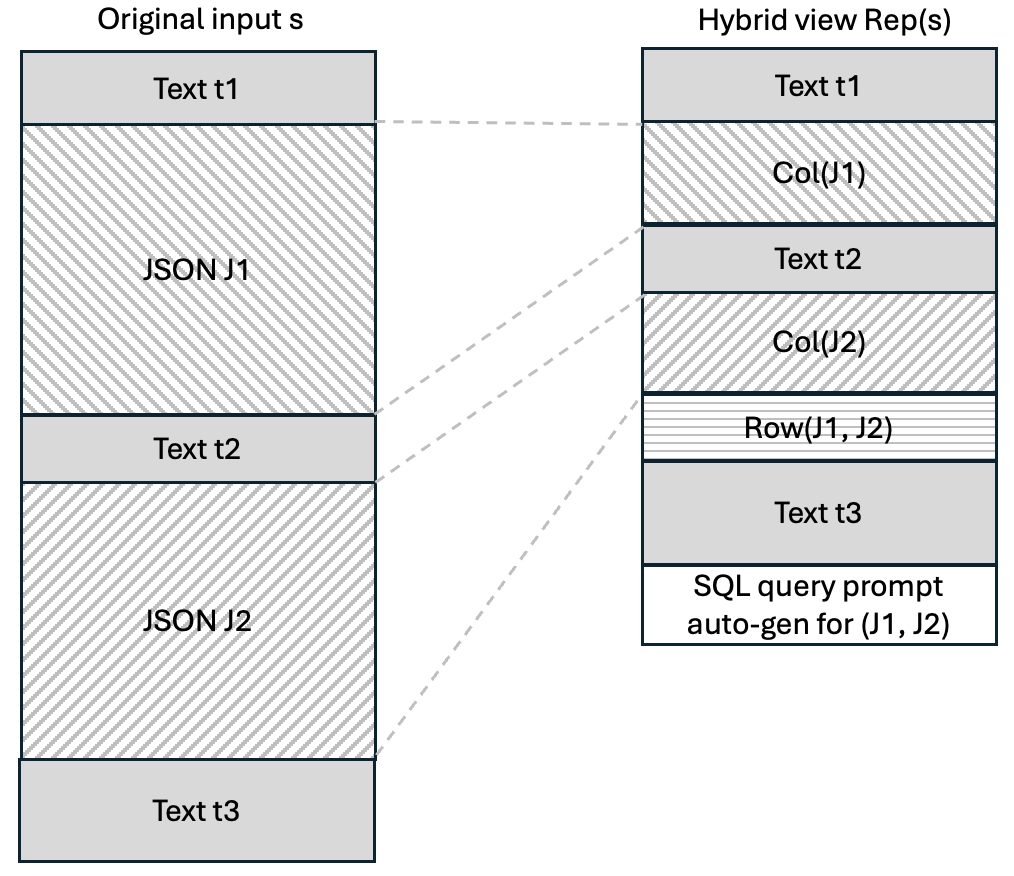}
\caption{Layout of $\mathsf{Rep}(s)$. If table truncation is triggered, SQL query prompt is automatically generated.
}
\label{fig:rep}
\end{figure}

\paragraph{Hybrid View}
For each object $J_i$, the preprocessor generates an ordered column view $\mathsf{Col}(J_i)$. The sequence $\mathsf{Col}(J_1), \mathsf{Col}(J_2), \ldots$ is interleaved with the text segments $\mathbf{t}$ in original order. It also generates a combined, unordered row view $\mathsf{Row}(\mathbf{J})$ across all extracted objects. As illustrated in Figure~\ref{fig:rep}, the resulting prompt $\mathsf{Rep}(s)$ is assembled as the following \emph{hybrid view}:
\begin{align}
t_1 \oplus \mathsf{Col}(J_1) \oplus t_2 \oplus \mathsf{Col}(J_2) \oplus \cdots \oplus t_{m+1} \oplus \mathsf{Row}(\mathbf{J}),\nonumber 
\end{align}
where $\oplus$ denotes string concatenation.

To make these two views precise, we define them formally below. The column view groups schema-aligned value sequences for analytics-oriented operations, whereas the row view preserves complete records for entity-level correlation.

 \begin{definition}[Column-Oriented Representation]
  \label{def:columnstore}
  Given a JSON object $J$, let $\mathcal{S}(J)$ be the set of scalar-value occurrences
  in $J$. Each occurrence is represented as a tuple $(c, p, v)$, where $v$ is the
  scalar value, $p$ is its flattened field path relative to its enclosing repeated
  record, and $c$ is its repeated-context path, defined as the ordered sequence of
  array-valued ancestor paths from the root of $J$ to that enclosing record.

  The column-oriented representation of $J$ is
  \[
  \mathsf{Col}(J)=\{((c_i,p_i), V_i)\}_{i=1}^{m},
  \]
  where each $(c_i,p_i)$ is a column identifier, and
  $V_i = [v_1,\ldots,v_{n_i}]$ is the ordered list of all scalar values whose
  occurrences in $\mathcal{S}(J)$ have the same repeated-context path $c_i$ and the
  same relative flattened field path $p_i$. The ordering of $V_i$ follows the
  original traversal order of the corresponding occurrences in $J$. Thus, two values are placed in the same column iff they are from schema-consistent repeated records, share the same repeated-context path, and have the same relative
    flattened field path.
  \end{definition}

  A simple example clarifies this. Consider the path \texttt{clusters[].nodes[].metrics[].cpu} and a corresponding one \texttt{clusters[].nodes[].metrics[].mem}. 
  They both have the same repeated-context path
  \texttt{clusters[].nodes[].metrics[]}, but with different relative flattened field paths, i.e., \texttt{cpu} and \texttt{mem}, respectively. Therefore,  their column identifiers can be represented as tuples
  $(\texttt{clusters[].nodes[].metrics[]}, \texttt{cpu})$ and $(\texttt{clusters[].nodes[].metrics[]}, \texttt{mem})$. By contrast, for a path
  \texttt{clusters[].nodes[].status}, the enclosing repeated record is an element of \texttt{nodes[]}, so its column identifier is
  $(\texttt{clusters[].nodes[]}, \texttt{status})$.

We transform nested JSON objects into a column-oriented representation suitable for analytics operators. The key idea is to distinguish \emph{schema-consistent} lists, whose elements share the same recursive schema signature as defined in Definition~\ref{def:consistent}, from \emph{schema-inconsistent} lists, whose elements differ in field structure or compatible leaf types. In the implementation, this check is performed bottom-up over the JSON tree (Section~\ref{sec:system}). This distinction enables safe columnar merging without conflating incompatible records.

  \begin{definition}[Row-Oriented Representation]
  \label{def:rowview}
  Given a collection of JSON objects $\mathbf{J} = (J_1,\ldots,J_m)$, its row-oriented
  representation is a set of records
  \[
  \mathsf{Row}(\mathbf{J})=\{r_1,\ldots,r_n\},
  \]
  where each $r_i$ is a complete schema-consistent repeated record extracted from
  some $J_k$. Each row corresponds to one original occurrence of that record in the
  input and retains all of its fields together as a single tuple, preserving the
  co-occurrence relationships among those fields. The row view may aggregate records
  from multiple objects and rank them by relevance to the request context; unlike
  the column view, it does not preserve the original global ordering of records.
  \end{definition}

  Consider the following example for illustration. If \texttt{clusters[].nodes[]} is a schema-consistent repeated record, then one row could be
  (\texttt{id}=\texttt{"node-1"},\ \texttt{status}=\texttt{"healthy"},\ \texttt{zone}=\texttt{"us-west"})
  and another row could be
  (\texttt{id}=\texttt{"node-2"},\ \texttt{status}=\texttt{"degraded"},\ \texttt{zone}=\texttt{"us-east"}).
  Each row keeps the fields of a single original node record together, so correlations such as which \texttt{status} belongs to which
  \texttt{id} remain explicit.

These two representations are complementary. The column view groups homogeneous values to support extraction and numerical analysis across fields, while the row view preserves record boundaries so that correlations among fields remain explicit. Together, they address the reasoning failures highlighted in Section~\ref{sec:intro}.

\paragraph{Table Entry Truncation}
The challenge is that both $\mathsf{Col}(J_i)$ and $\mathsf{Row}(\mathbf{J})$ may still be too large to expose in full. HYVE therefore truncates only \emph{table entries} while preserving schema. For the column view, the retained subset must preserve the original order to maintain semantic continuity. For the row view, ordering is not needed, because its primary purpose is to expose representative entities and their relationships.

HYVE uses reference-aware term-weighting methods from information retrieval to decide which entries in $\mathsf{Col}(J_i)$ and $\mathsf{Row}(\mathbf{J})$ are most relevant to the task. To do so, it constructs a \emph{query string} $q$ by concatenating the surrounding text segments $t_1, \ldots, t_{m+1}$, which typically contain the user’s question or instruction, with a slim representation $\mathbf{Slim}(\mathbf{J})$ that retains all key-value pairs outside the detected tables. The role of $\mathbf{Slim}(\mathbf{J})$ is important: non-tabular key-value pairs often encode request-level semantics that are needed to rank table entries correctly.

Specifically, HYVE truncates table entries as follows:
  \begin{itemize}
  \item $\mathsf{Col}(J_i)$: Long columns are truncated to a fixed-length prefix and postfix, together with additional items selected by a
  reference-aware ranking score with respect to $q$, while preserving the original order.
  \item $\mathsf{Row}(\mathbf{J})$: Only a subset of complete records is presented, ranked by the same method to highlight entities and their correlations
  without preserving the original order.
  \end{itemize}

To preserve access to the full dataset, the preprocessor builds a \emph{datastore} $\mathcal{D}(\mathbf{J})$, in which the untruncated data are loaded into in-memory relational tables with schemas inferred from the input, including column names, value types, and parent-child foreign keys for nested arrays.  Specifically, the repeated-context path $c$ helps determine the relational table to which a value belongs, while the flattened field path $p$ determines
  the corresponding field name; nested arrays are normalized into parent-child tables.

\subsection{Datastore and SQL Query}

When table entry truncation is triggered, 
HYVE automatically injects tool-call prompts to guide SQL generation.

\paragraph{SQL Prompt Auto-Generation}
\label{par:sql}
 
Reliable SQL generation remains challenging for LLMs in open-ended settings, but our environment is controlled and provides exact knowledge of the schema and parent-child relationships among tables.
  HYVE synthesizes a schema-aware tool prompt from $\mathcal{D}(\mathcal{J})$, listing the available tables, columns, parent-child keys, and the output contract of the permitted tool call. It also constructs targeted few-shot examples to guide the LLM toward generating high-quality SQL queries for each request.

 Because these prompts are generated per request and may include corner-case details, HYVE adopts a gradual-disclosure design. In the primary LLM call,
  the model only needs to identify the task intent and choose the appropriate tool. When a follow-up call is needed, HYVE provides more contextualized
  prompts to guide SQL generation and data backfilling. Thus, HYVE incurs at most one additional LLM call by default. 

In practice, such guidance can be supplied, for example, through the \texttt{instructions} field in the OpenAI Responses API~\cite{openai_responses_api} or the Chat Completions API~\cite{openai_chat_completions_api}; alternatively, the same design can be implemented through Anthropic’s API or its OpenAI SDK compatibility layer~\cite{anthropic_messages_api,anthropic_openai_sdk_compatibility}.

We illustrate this process with a real example. The following snippet shows $\mathsf{Rep}(s)$ after preprocessing the input $s$. For clarity, 
we retain only the first three data entries in the column view and omit non-essential descriptions. 
The transformed prompt $\pi$ exposes only a small visible slice of the original chart while injecting 
relational metadata for SQL generation. In this example, the user asks for a multi-series line chart over path-performance telemetry, 
and the preprocessor retains truncated column views, presents a short row-view sample, and then appends 
the tool-selection prompt and datastore schema:
\begin{lstlisting}[basicstyle=\ttfamily\scriptsize,breaklines=true]
<language descriptions/instructions omitted...>
  "(CARD 1)": {
    "name": "CDSAILineChart",
    "props": {
      "data": {
        "id": [
          "Newark, NJ", "pdx-linux-ea", "New York, NY"
        ],
        "data": {
          "x": [
            [1770163200000,1770163500000,1770163800000],
            [1770163200000,1770163500000,1770163800000],
            [1770163200000,1770163500000,1770163800000]
          ],
          "y": [
            [2.1, 1.8, 2.2],
            [17.7, 18.0, 16.0],
            [28.1, 26.3, 22.9]
          ]
        }
      },
      "axes": {
        "xAxis": {
          "type": "time"
        }
      }
    }
  }
<other instructions/intents/questions omitted...>
### Row View:
    data:
      x | y_0 | y_1 | y_2
      ------------------
      1770163200000 | 2.1 | 17.7 | 28.1
      1770163500000 | 1.8 | 18.0 | 26.3
      1770163800000 | 2.2 | 16.0 | 22.9
      1770164100000 | 2.1 | 15.3 | 23.1
      1770164400000 | 2.4 | 16.1 | 24.7
      1770164700000 | 2.0 | 15.9 | 24.0
      ... (762 more rows)

###Tool Selection (data is truncated, with the complete data in datastore):
<tool call specifications omitted...>
Available tables in datastore:
Table: data
  Source: CARD 1.props.data.data
  Total rows: 2304
  Rows per series: 768
  series_idx values: 0='Newark, NJ', 1='pdx-linux-ea', 2='New York, NY'
  Columns:
    - series_idx (int)
    - x (int)
    - y (float)
    - _row_id (int)
### Few-Shot Query Examples:
<auto-gen to guide LLM SQL generation; details omitted...>
\end{lstlisting}

This excerpt clarifies the role of $\pi$: it is not merely a compressed serialization of the raw input, but a hybrid prompt that combines visible
  evidence with truncation metadata. When data-entry truncation is triggered, additional SQL-generation prompts, table-schema information, and few-shot
  guidance are also provided. The LLM therefore reasons over the visible slice while using the injected table description to decide whether it can answer
  directly, should call \texttt{QueryDatastore}, or should initiate \texttt{GenTemplateAndBackfill}.

\subsection{LLM Call and Postprocessor}
If the visible prompt is insufficient for the primary LLM call, HYVE completes the result through either template-based data backfilling or query-based evidence augmentation.
\paragraph{LLM Output}
Given a prompt $\pi$, the LLM generates $\mathsf{LLM}(\pi)\rightarrow y$, 
where $y$ may be either a free-form answer or a structured object conforming to a schema, 
such as a tool-call output or a visualization specification. 

Below is an illustrative output for the preceding example, produced in response to \texttt{GenTemplateAndBackfill}. It contains an answer template derived from the visible data together with a \texttt{BackfillData} tool call. This tool call includes SQL statements and mappings that specify how the query results should be backfilled into the template. By default, this process uses one additional SQL query.

   \begin{lstlisting}[
    language=json,
    title={\textit{The answer template generated by the LLM}},
    basicstyle=\ttfamily\scriptsize,
    breaklines=true,
    keepspaces=true,
    columns=fullflexible,
    xleftmargin=1em,
    aboveskip=4pt,
    belowskip=4pt
  ]
  {
    "CDSAILineChart": {
      "data": {
        "list": [
          {
            "series_name": "Newark, NJ",
            "data": [
              {
                "x": 1770163200000,
                "y": 2.1
              }
              // ... will be filled in for Newark, NJ
            ]
          },
          {
            "series_name": "pdx-linux-ea",
            "data": [
              // ... will be filled in for pdx-linux-ea
            ]
          },
          {
            "series_name": "New York, NY",
            "data": [
              // ... will be filled in for New York, NY
            ]
          }
        ]
      },
      "axes": {
        "xAxis": {
          "type": "time"
        }
      }
    }
  }
  \end{lstlisting}

\begin{lstlisting}[
    language=json,
    title={\textit{The SQL tool-call specification}},
    basicstyle=\ttfamily\scriptsize,
    breaklines=true,
    keepspaces=true,
    columns=fullflexible,
    xleftmargin=1em,
    aboveskip=4pt,
    belowskip=4pt
  ]
  {
    "tool_name": "BackfillData",
    "query": "SELECT x, y, series_idx FROM data ORDER BY series_idx, _row_id",
    "mappings": [
      {
        "sql_column": "x",
        "template_path": "CDSAILineChart.data.list.(series_idx).data.(_row_id).x"
      },
      {
        "sql_column": "y",
        "template_path": "CDSAILineChart.data.list.(series_idx).data.(_row_id).y"
      }
    ]
  }
  \end{lstlisting}
This output requests a \texttt{BackfillData} operation over the full datastore. The SQL query retrieves the complete \texttt{x} and 
\texttt{y} series together with \texttt{series\_idx} and \texttt{\_row\_id}, ordered so that each series can be reconstructed in its original order. 
Each mapping specifies how one SQL column should be written back into the output template. 
Here, \texttt{(series\_idx)} identifies the target series, 
while the final indexed position \texttt{(\_row\_id)} denotes 
the element within that series. The postprocessor can therefore 
reconstruct the full chart data while preserving the schema produced by the LLM.

 \paragraph{Postprocessor Working Modes}
  If preprocessing does not trigger truncation, the LLM operates directly on $\pi$ and produces the final result $y$; no datastore access or postprocessing
  is required. If truncation does occur, however, part of the request-side evidence is no longer visible to the LLM. HYVE therefore synthesizes a schema-aware SQL prompt from $\mathcal{D}$, enabling the model to emit SQL statements that are executed by the postprocessor. This yields three operating modes:
  \begin{itemize}
  \item \emph{Mode~1 (Direct Output).} When the visible context is sufficient, for example for a point lookup, the LLM output $y$ is returned directly.
  \item \emph{Mode~2 (Template Backfill).} This mode corresponds to the \texttt{GenTemplateAndBackfill} branch introduced above and targets rendering tasks
  in which the LLM first generates the schema of a structured answer, typically under a JSON specification. HYVE uses gradual disclosure in this branch:
  the primary prompt presents \texttt{GenTemplateAndBackfill} only as a routing choice, while the detailed \texttt{BackfillData} tool contract is revealed
  only after Mode~2 has been selected. The follow-up prompt then instructs the LLM to produce both the answer template and a \texttt{BackfillData} tool
  call containing SQL queries together with column-to-path mappings. The postprocessor executes the query against $\mathcal{D}$ and injects the full result
  into the template according to those mappings. In this way, the visible portion is generated by the LLM, while the hidden portion is restored through
  template-based expansion that preserves the model-generated schema.
  \item \emph{Mode~3 (SQL-Augmented Synthesis).} This mode corresponds to the \texttt{QueryDatastore} branch introduced above. Unlike Mode~2, the primary
  prompt already includes the full \texttt{QueryDatastore} tool specification, so no second tool-disclosure step is required. When the task requires
  semantic synthesis over both the queried data and the natural-language prompt, such as aggregation, counting, filtering, or joins across multiple tables,
  the primary LLM emits a \texttt{QueryDatastore} tool call containing one or more SQL queries. The postprocessor executes these queries against $
  \mathcal{D}$, appends the formatted results to the conversation context, and makes one bounded additional LLM call to produce a synthesized answer
  grounded in the complete evidence.
  \end{itemize}
Under the default design, HYVE requires only the primary LLM call in Mode~1 and at most one additional LLM call in Modes~2 and~3. As discussed in
Appendix~\ref{sec:bounded}, Mode~3 can also be configured to allow a bounded number of additional calls to improve the fidelity of the final answer. The
  previous example operates in Mode~2.

\section{System Architecture and Implementation}
\label{sec:system}
This section presents the system architecture and the main implementation details.
\subsection{System design}
At a high level, HYVE consists of three coordinated components (Figure~\ref{fig:architecture}) 
that operate within an isolated per-request scope.
  Each request is fully isolated, and the lifecycles of all components are confined to that scope. 
  To support an in-memory datastore and expressive
  SQL features (e.g., extraction operators for nested JSON objects), 
  HYVE uses DuckDB~\cite{duckdb2019}. Concretely, it instantiates an unnamed in-memory
  DuckDB database for each request via the special \texttt{:memory:} target. This yields an efficient transient datastore. No data are written to
  disk, the database is discarded when the request finishes, and separate unnamed \texttt{:memory:} connections prevent cross-request state sharing
  by construction.

HYVE can be integrated with standard LLM APIs in a provider-agnostic manner, for example through OpenAI-style interfaces such as the Responses API~\cite{openai_responses_api} or
  Chat Completions API~\cite{openai_chat_completions_api}, as well as Anthropic’s Messages API~\cite{anthropic_messages_api}. HYVE
  remains responsible for structured parsing, datastore construction, truncation-aware representation, and query-based output repair, while the underlying
  LLM API handles request execution and tool-calling transport.

\subsection{Preprocessor Implementation}
Given a raw input string, HYVE extracts structured text segments and parses each into a canonical object. It materializes these objects in the datastore before any truncation. HYVE then builds the column view by constructing a JSON tree, flattening it into JSONPath-like dotted keys, and
  clustering paths that share the same repetitive structural pattern after abstracting away index values. These clusters are merged into schema-aware columns whose values are reshaped into ordered lists. When lists are large, HYVE applies truncation and ranking so that high-signal table entries
  can still appear in the visible prompt. HYVE similarly derives the row view by ranking records against the request context, selecting a small
  representative subset, and rendering the result as a compact pipe-delimited table. Algorithm~\ref{alg:preprocess-hybrid-view} summarizes this workflow.

\begin{algorithm}[t]
\caption{Pre-processing and Hybrid-View Construction}
\label{alg:preprocess-hybrid-view}
\begin{algorithmic}[1]
\STATE \textbf{Input:} Raw input string $s$, query text $q$ for ranking
\STATE \textbf{Output:} Prompt-ready hybrid view and full datastore
\STATE $(T, O) \gets \textsc{ExtractTextAndJSONObjects}(s)$
\STATE $\mathcal{D} \gets \textsc{NewDatastore}()$
\STATE $\mathcal{C} \gets [\ ]$
\FOR{each object $o \in O$}
    \STATE $J_o \gets \textsc{Parse}(o)$
    \STATE $\textsc{BuildDatastoreFromObject}(\mathcal{D}, J_o)$
    \STATE $\tau \gets \textsc{BuildJSONTree}(J_o)$
    \STATE $\textsc{NormalizeExpandedUnparsedNodes}(\tau)$
    \STATE $\textsc{MarkSchemaInconsistency}(\tau)$
    \STATE $C \gets \textsc{ClusterByJSONPathPattern}(\tau)$
    \STATE $E \gets \textsc{MergeAndReshapeColumns}(C)$
    \STATE $R \gets \textsc{RankSchemaCompatibleIndexes}(E, q)$
    \STATE $E' \gets \textsc{ApplyTruncationAndRanking}(E, R)$
    \STATE $\mathcal{C}.\textsc{Append}(\textsc{RenderColumnView}(E'))$
\ENDFOR
\STATE $\mathsf{Row}(\mathbf{J}) \gets \textsc{RenderRankedRows}(\mathcal{D}, q)$
\STATE $\pi \gets \textsc{AssemblePrompt}(T, \mathcal{C}, \mathsf{Row}(\mathbf{J}))$
\RETURN $(\pi, \mathcal{D})$
\end{algorithmic}
\end{algorithm}

\paragraph{JSON Tree and Schema Consistency}
The JSON tree is the intermediate representation that makes this transformation precise. Each node stores a key, a value, whether it originated from a dictionary or a list, and links to its children. Dictionary edges preserve semantic field names, whereas list edges are represented by numeric child keys. During tree construction, HYVE also attempts to expand long string values that themselves contain JSON or Python-literal payloads, so that nested structure is exposed rather than treated as opaque text. Flattening the tree then yields dotted paths such as \texttt{data.0.x}, \texttt{data.1.x}, and \texttt{data.2.x}. Clustering collapses these paths to a shared pattern such as \texttt{data.*.x}, while retaining capture cardinalities so that the collected values can be reshaped into aligned lists.

Schema consistency is required because clustering is valid only when different list items instantiate the same record schema. 
HYVE therefore performs a bottom-up check over every list node. For each list item, it computes a \emph{deep schema signature} 
that recursively records child keys and normalized leaf types, treating \texttt{int} and \texttt{float} uniformly as \texttt{number}.
 A list is declared schema-consistent only if all items share the same signature and no descendant leaf exceeds a configured maximum length for column-view rendering. This check prevents false alignment. Without it, unrelated objects such as \texttt{\{x,y\}} and \texttt{\{name,status,error\}} could be merged into the same column family simply because they occupy the same list position, producing malformed columns and invalid row correspondences.

When this check fails, HYVE marks the list node as schema-inconsistent and preserves the numeric indices of all list items explicitly. This disables clustering across those items and ensures that only entries with the same inferred schema are merged
  into a single logical table or column family. The same safeguard applies to enumerated dictionary children when they contain sufficiently long payloads, as
  determined by a configuration parameter.
   HYVE preserves their indices rather than forcing them into a shared columnar pattern. 
   As a result, schema-consistent regions are compressed into tables, whereas schema-inconsistent regions remain as indexed records.

\paragraph{Relationalization of Nested Lists}
Datastore construction follows a complementary invariant. HYVE recursively scans the parsed object and promotes sufficiently large lists of dictionaries to parent tables. Inside each parent row, list-valued fields are split into child tables using foreign keys of the form \texttt{parent\_<id>}; if those child rows still contain nested lists, the same rule is applied recursively to produce grandchild tables. Flat one-to-one dictionaries are expanded into columns of the current table, while residual nested objects are serialized as JSON strings when they cannot be normalized safely. The implementation also handles two important multi-series cases. First, for column-store dictionaries whose values are parallel lists of lists, HYVE converts them into a single flat table with a \texttt{series\_idx} column and attaches sibling label arrays as series metadata. Second, when a list of sibling objects contains parallel sublists under the same key, HYVE combines those sublists into one shared table, again indexed by \texttt{series\_idx}. These rules ensure that lists at different depths are either mapped to legitimate relational tables with explicit join keys or preserved as indexed nested structures when no sound relational alignment exists.

\paragraph{Robust Parsing}
Given the raw input string $s$, the preprocessor first identifies candidate structured segments as part of the $\mathsf{Parse}$ operator (Section~\ref{sec:problem}). Each candidate segment is then passed to a multi-strategy robust parser that attempts to recover a canonical structured object. This parser is designed to handle malformed JSON, string-encoded nested objects, and Python AST literals.

\begin{algorithm}[t]
\caption{Robust Parsing of a Candidate Segment}
\label{alg:robust-parsing}
\begin{algorithmic}[1]
\STATE \textbf{Input:} Candidate structured segment $s$
\STATE \textbf{Output:} Parsed structured object $O$ or the original segment $s$
\STATE $s_0 \gets \textsc{StripMarkdownFences}(s)$
\STATE \textbf{Stage 1: Fast parsing}
\STATE Try, in order: \textsc{ParseCanonicalJSON}, \textsc{ParsePythonLiteral}
\FOR{each parser $P$ in the above order}
  \STATE $O \gets P(s_0)$
  \IF{$O \neq failure$}
        \STATE $O \gets \textsc{ExpandNestedJSON}(O)$
    \IF{$\textsc{ValidateSourceCoverage}(O, s_0)$}
      \RETURN $O$
    \ENDIF
  \ENDIF
\ENDFOR
\STATE $s_1 \gets \textsc{NormalizeControlCharacters}(s_0)$
\STATE $s_1 \gets \textsc{NormalizeKeyQuoting}(s_1)$
\STATE $s_1 \gets \textsc{StabilizeNestedJSONStringSyntax}(s_1)$
\STATE \textbf{Stage 2: Repair-oriented recovery}
\STATE Try, in order: \textsc{RepairMalformedJSON}, \textsc{DecodeEncodedJSON}, \textsc{RecoverEmbeddedJSONSpan}
\FOR{each repairer $R$ in the above order}
  \STATE $O \gets R(s_1)$
  \IF{$O \neq failure$}
    \STATE $\textsc{ReconcileWithOriginal}(O, s_0)$
    \STATE $\textsc{PreserveUnrecoveredContent}(O, s_0)$
    \STATE $O \gets \textsc{ExpandNestedJSON}(O)$
    \IF{$\textsc{ValidateSourceCoverage}(O, s_0)$}
      \RETURN $O$
    \ENDIF
  \ENDIF
\ENDFOR
\RETURN $s$
\end{algorithmic}
\end{algorithm}

The parser follows a two-stage design. The \emph{fast parsing stage} targets common cases with low overhead. It first strips surrounding Markdown code fences, then tries a small sequence of lightweight parsers that move from strict to more permissive formats: canonical JSON parsing, followed by Python-literal parsing with only minimal syntax normalization needed to expose the literal itself. This stage does not attempt broad repair. Whenever one of these parsers succeeds, HYVE recursively traverses the resulting dict/list structure. If a string-valued field itself parses as a JSON object or array, HYVE replaces that string with the parsed structure and continues the traversal on the newly exposed subtree until no further expansion applies. This recursive expansion is necessary because real inputs often contain nested payloads that are string-encoded inside outer JSON fields; without it, deeper records would remain opaque text and could not be normalized into the hybrid view or datastore. This stage covers many practical inputs, including code-fenced JSON, Python dict/list literals, and stringified nested payloads.

If the fast stage fails, HYVE enters a \emph{repair stage}. Before invoking more permissive recovery, the parser applies targeted normalization to make the candidate structurally coherent: it normalizes control characters that would invalidate JSON strings, inserts quotes around unquoted field names, and escapes quotes inside nested JSON fragments embedded within string values. HYVE then proceeds through a small sequence of increasingly permissive recovery strategies. It first attempts structural repair of malformed JSON. If that still fails, it handles wrapper cases in which the payload is itself encoded as a quoted or escaped JSON string. Finally, when the candidate contains surrounding free-form text, HYVE extracts the most plausible embedded JSON span and, when necessary, restores missing closing or opening braces/brackets so that the recovered span forms a complete object or array. Because permissive recovery may silently drop or rewrite malformed content, each recovered candidate is reconciled against the original segment: content that is confidently recovered is kept in structured form, while any unrecovered residue is retained under a dedicated \texttt{unparsed\_string} field rather than discarded. Finally, each candidate parse is validated against the original input by measuring overlap in alphabetic-word substrings.
HYVE rejects parses that violate the word-overlap coverage criterion, 
thereby reducing silent truncation. Algorithm~\ref{alg:robust-parsing} summarizes this
  control flow.

\section{Operator Design}
\label{sec:preprocessing}
HYVE includes both generic and specialized operators, including ranking, rendering, and time-series analysis.

\subsection{Ranking-Based Subset Selection}
  \label{sec:ranking}

  HYVE uses ranking in two settings. The row view supports \emph{retrieval-oriented} tasks, where the goal is to select the most relevant records from a large collection. The column view preserves a prefix and suffix together with a small number of additional elements selected by a ranking function, while maintaining the original order.

  HYVE uses two related ranking procedures with different candidate corpora. For the column view, ranking is performed independently within
  each \emph{schema-compatible} group. Each list position defines one candidate record, and ranking statistics are computed using only the
  records in that group. Because all records in the same group share the same field paths, schema tokens contribute no discriminative signal.
  Accordingly, column-view ranking uses only value tokens.
 
  For the row view, the candidate corpus is the union of candidate rows from the request-scoped datastore.
  Here, each candidate is represented by both schema tokens and value tokens. Value tokens provide the fine-grained matching signal, while
  schema tokens identify the table and field family from which the values originate. This distinction matters when the same request contains
  multiple heterogeneous tables or repeated structures whose values may overlap lexically. 

  We score these candidates with a standard BM25-style ranker~\cite{robertson2009probabilistic}. The query is a reference string formed from the
  surrounding input context together with $\mathbf{Slim}(\mathbf{J})$. The key point is not the scoring function itself, but how HYVE maps structured data into retrieval units. BM25 is applied not to free-form documents, but to structured
  records. For column-view ranking, IDF is computed within each schema-compatible group; for row-view ranking, it is computed over the
  cross-table candidate set. Token overlap with the reference context determines relevance, and repeated occurrences in the reference are
  saturated in the usual BM25 manner.

  Relative to standard document retrieval, HYVE differs in two respects. First, its retrieval units are structured records augmented with schema
  information rather than free-form documents. Second, ranking is performed against a reference string derived from the request context. HYVE therefore
  applies BM25 in a structured setting, matching request-derived context to candidate records rather than ranking free-text documents.


\subsection{Time-Series Analysis Operators}
\label{sec:anomaly}

For structured inputs containing numerical time series or metric sequences, we provide optional statistical and anomaly-detection operators that convert raw numerical arrays into compact, high-signal natural-language summaries.

Long metric sequences, such as per-node network delays collected along a path, can consume substantial tokens while hiding the few data points that matter most. Blind truncation may discard the rare high-latency events that determine the diagnosis. HYVE therefore supports time-series analysis operators that compress such sequences into compact, semantically meaningful evidence, such as anomaly reports and trend descriptions.

As a simple illustration, we implement an $n$-sigma anomaly detector. Given a numerical sequence $X = [x_1, \ldots, x_n]$, it flags observations whose
  deviation from the mean exceeds a fixed multiple of the standard deviation (with default threshold $k = 2$), while applying a lightweight periodicity
  check to suppress false positives~\cite{chandola2009anomaly}. We expose this analysis as a query operator over the datastore. For instance, an
  LLM can invoke a custom SQL-style function, \texttt{DETECT\_ANOMALY(data, y)}, to identify abnormal behavior in metric column \texttt{y} of table
  \texttt{data}. Rather than returning the full raw time series, the operator produces compact textual evidence.

  For example, it may return a concise summary such as:
\begin{lstlisting}[basicstyle=\ttfamily\footnotesize]
    Anomaly detected in data.y:
      latency: 293.0 ms 
    Normal observation: 
      mean: 3.08 ms, max: 11.19 ms
    Context:
      timestamp: 2025-07-15T10:30:00
      pathTrace: 4, round: 9, hop: 4
      node: "pdx-linux-ea"
\end{lstlisting}
This summary surfaces the anomalous value together with relevant contextual fields and a brief statistical characterization of normal behavior. By
  transforming raw numerical sequences into compact, semantically meaningful summaries, such operators help the LLM focus on the most salient evidence
  without being overwhelmed by token-intensive raw data.

\subsection{Rendering Operators}
\label{sec:rendering}

After flattening and truncation, we reconstruct a serialized representation suitable for the LLM prompt. 
This rendering step converts the internal column-oriented representation into a human-readable, LLM-consumable format.

We explored multiple output formats:
  \begin{itemize}
  \item \textbf{Beautified JSON (default):} Reconstruct the nested JSON object with consistent indentation for readability.
  \item \textbf{Raw JSON:} Preserve the JSON in minified form, eliminating unnecessary whitespace to maximize token density at the expense of readability.
  \item \textbf{TOON encoding:} Use a compact, lossless representation of the JSON data model for LLM input~\cite{toon}, combining YAML-like indentation
  with CSV-style tabular layouts for uniform arrays of objects.
  \end{itemize}

TOON encoding offers strong token efficiency for table-like structured data. Our ablation study (Section~\ref{sec:ablation}) shows that it yields up to 31\% token savings, with the largest gains on chart generation and slot-filling tasks (18--31\%), though savings are smaller (6--15\%) or negligible on other workloads. However, TOON can degrade structured-generation quality. We therefore use beautified JSON as the default rendering format, prioritizing
  broad task compatibility over maximum token savings. TOON remains a viable option when stronger token efficiency is desired.

  As work orthogonal to the main theme of this paper, we also introduce operators for other input formats. For example, an HTML operator can transform
  HTML-formatted text into a browser-like rendered view, reducing formatting metadata that might otherwise obscure the key information for the LLM.

\section{Experimental Evaluation}
\label{sec:eval}

We evaluate HYVE on a diverse collection of real-world networking datasets involving machine data and structured outputs. Most of these
  datasets are proprietary, so we describe them in the appendix. We also evaluate HYVE on the public TOON-QA benchmark~\cite{toon}.

\subsection{Datasets}
\label{sec:datasets}

The datasets include expert-authored and workflow-oriented benchmarks spanning question answering, slot filling, 
anomaly detection, chart generation, and network troubleshooting.
We group our benchmarks into the following categories:
\begin{itemize}
\item \textbf{Cert-QA:} A collection of 3,524 network engineering questions spanning Cisco certification levels.
This includes CCNA-level entry questions on basic commands and protocols, CCNP-level questions on advanced routing (OSPF, BGP) and switching (STP, MST), and CCIE-level expert questions covering unified communications, wireless security (802.1X, WPA2/3), and network analytics.
The dataset combines both multiple-choice exam questions with detailed feedback and open-ended questions requiring precise technical answers.
Representative examples are provided in Appendix~\ref{app:cert-qa-examples}.
\item \textbf{Runbook:} A collection of 260 network-operations troubleshooting scenarios derived from real-world support workflows. Each example contains a problem
  description, such as a stack upgrade failure or port connectivity issue, together with a ground-truth runbook specifying step-by-step
  diagnosis and verification procedures. The task requires the model to generate or complete a structured troubleshooting playbook from the
  problem context.
A representative example is provided in Appendix~\ref{app:runbook-examples}.
\item \textbf{Line:} 100 line-chart generation tasks that require the model to transform time-series network telemetry, such as latency, jitter, packet loss, and
  goodput, into multi-series line-chart JSON specifications with appropriate axis configurations and data mappings.
A representative example is provided in Appendix~\ref{app:line-chart-examples}.
\item \textbf{Bar:} 100 bar chart generation tasks requiring the model to convert categorical key-value data into bar-chart JSON schemas with stacks and category labels.
A representative example is provided in Appendix~\ref{app:bar-chart-examples}.
\item \textbf{Anom:} 797 anomaly detection tasks requiring the model to analyze network path visibility data and identify high-latency nodes.
Each entry contains hop-by-hop response times from synthetic network tests, and the model must apply a latency threshold (e.g., $>$10ms) to flag anomalous nodes, then provide structured reasoning and a summary of impacted hosts.
A representative example is provided in Appendix~\ref{app:anom-examples}.
\item \textbf{Sum:} 84 report generation tasks requiring the model to synthesize structured board/canvas data into comprehensive Markdown reports.
Each entry contains a JSON object with board metadata, canvas details, cards, and conversation history; the model must produce a well-organized report with a table of contents, key insights, incident timelines, and root cause analysis where applicable.
A representative example is provided in Appendix~\ref{app:sum-examples}.
\item \textbf{Canvas:} 4,096 slot-filling tasks aggregated from 8 runbook step scenarios.
Each entry simulates a CCIE-level network troubleshooting workflow where the model receives a deeply nested board context (metadata, canvas, cards, conversation history) along with a runbook step instruction, and must extract or compute the required variables (e.g., \texttt{orgId}, \texttt{networkId}, time ranges, API query parameters) to proceed.
A representative example is provided in Appendix~\ref{app:canvas-examples}.
\item \textbf{RB-Text:} A collection of API response summarization tasks derived from network troubleshooting runbook executions.
Each entry contains a raw API response from commercial network management and monitoring platforms (e.g., device alerts, monitored targets, connectivity status) along with a CCIE-expert system prompt; the model must produce a concise, well-structured Markdown summary that highlights critical information such as alert severity, affected devices, and actionable insights.
A representative example is provided in Appendix~\ref{app:rb-text-examples}.
\item \textbf{RB-JSON:} Representative conditional expression evaluation tasks requiring the model to 
assess boolean predicates over runbook execution context.
Each entry presents a condition (e.g., ``Location is set'', ``Fault Domain is local-network'', ``Any of the alerts are critical'') along with structured context data from network troubleshooting workflows; the model must output a JSON object containing a step-by-step reasoning array and a boolean decision indicating whether the condition holds.
A representative example is provided in Appendix~\ref{app:rb-json-examples}.
\item \textbf{Hard:} A curated set of challenging multi-hop reasoning problems designed to stress-test structured data comprehension.
Each entry presents a complex query (e.g., ``find a DNS test related to SharePoint running from an agent in San Francisco'') over deeply nested JSON payloads containing network test configurations, agent metadata, and location hierarchies.
The model must perform multi-step filtering and cross-referencing to produce a detailed reasoning chain and the correct answer.
A representative example is provided in Appendix~\ref{app:hard-examples}.
\item \textbf{TOON-QA:} A subset of 154 question-answer pairs over JSON data selected from the TOON Retrieval Accuracy Benchmark~\cite{toon}.
Questions span three categories: \emph{field retrieval} (direct value lookups, 25\%), \emph{aggregation} (counting, averaging, and statistical computation across records, 58\%), and \emph{multi-condition filtering} (compound queries requiring cross-field logic, 17\%).
Prompt lengths range from 10K to 43K characters (mean 26K), making this benchmark a stress test of LLM analytical reasoning over large structured contexts where the answer cannot be obtained by locating a single datum.
A representative example is provided in Appendix~\ref{app:toon-qa-examples}.
\end{itemize}

\subsection{Metrics}
\label{sec:metrics}

We evaluate system performance along three dimensions: \emph{answer quality}, \emph{token efficiency}, and \emph{latency}. All evaluations are conducted using an automated evaluation pipeline built on LangSmith~\cite{langsmith}.

\paragraph{Quality Metrics}
We employ task-appropriate quality metrics tailored to each dataset's characteristics:

\begin{itemize}
\item \textbf{GenericJudge (5-point scale):} An LLM-as-a-judge evaluator that compares the generated answer against a ground-truth reference. A GPT-4o judge assigns an integer score from 1 to 5 based on factual correctness, completeness, and clarity: \emph{5 (Excellent)}---fully correct and complete; \emph{4 (Good)}---mostly correct with minor omissions; \emph{3 (Fair)}---partially correct with notable gaps; \emph{2 (Poor)}---largely incorrect or incomplete; \emph{1 (Inadequate)}---completely incorrect or irrelevant. This metric is applied to \textbf{Cert-QA}, \textbf{Runbook}, \textbf{Sum}, and \textbf{RB-Text} datasets.

\item \textbf{Similarity (0--1 scale):} For chart generation tasks, we compute a composite similarity score between the model's output data series and the ground-truth series. The metric combines three components: (1) normalized Dynamic Time Warping (nDTW) similarity, which measures sequence alignment under temporal distortion; (2) cosine similarity, which captures directional agreement between value vectors; and (3) Pearson correlation, which measures linear relationship strength. The final score is the geometric mean of these three components (each normalized to $[0, 1]$), producing a holistic measure of data fidelity. This metric is applied to \textbf{Line} and \textbf{Bar} chart generation tasks, where the model must produce JSON specifications containing numerical data arrays.

\item \textbf{ReasoningJudge (5-point scale):} A specialized LLM-as-a-judge evaluator for structured JSON outputs. A GPT-4.1 judge first validates that the model's output is well-formed JSON conforming to a provided schema, then performs field-by-field comparison against the ground truth. String fields require exact matches; reasoning arrays allow semantic equivalence with paraphrasing permitted if all factual points are preserved. Schema violations cap the maximum score at 2. This metric is applied to \textbf{Anom} anomaly detection, \textbf{Canvas} slot-filling, \textbf{RB-JSON} conditional evaluation, and \textbf{Hard} multi-hop reasoning tasks, which require structured outputs with explicit reasoning chains.

\item \textbf{ExactMatch (0--1 scale):} A deterministic evaluator that compares the model's answer against the ground truth via exact string matching (case-insensitive, after whitespace trimming). A score of 1.0 is assigned if the answer matches exactly, and 0.0 otherwise. The final score is the mean accuracy across all examples. This metric requires no LLM judge and is fully reproducible. It is applied to \textbf{TOON-QA}, where answers are short deterministic values (integers, floats, or short strings) that admit unambiguous verification.
\end{itemize}

\noindent The complete evaluation prompts are provided in Appendix~\ref{app:eval-prompts}.

\paragraph{Token Usage}
We measure the \emph{total token consumption} for each method by summing the prompt and completion tokens across all examples in a dataset. Token counts are obtained directly from the model's API response. This metric captures the end-to-end cost of processing an entire benchmark, enabling direct comparison of encoding efficiency across methods.

\paragraph{Latency}
We report the \emph{mean end-to-end latency} per query, measured as the wall-clock time from request submission to response completion. For each dataset, we compute the average latency across all examples, providing a practical measure of user-perceived response time under each encoding strategy.

\subsection{Baselines}
\label{sec:baselines}

  We evaluate HYVE by applying it on top of two OpenAI baselines in their default configurations, i.e., without HYVE preprocessing or postprocessing:
  \begin{itemize}
  \item \textbf{GPT-4.1:} OpenAI’s GPT-4.1 model, a strong low-latency non-reasoning model for instruction following and coding, with standard JSON input
  serialization and no output recovery.
  \item \textbf{GPT-5:} OpenAI’s GPT-5 model, a newer general-purpose model designed for more complex tasks, under the same configuration.
  \end{itemize}
  These baselines reflect common practice: structured data are serialized directly as JSON and passed to the LLM. We compare performance with and without
  HYVE.

We additionally conduct ablation experiments to isolate the contribution of individual HYVE components (truncation strategies, rendering formats, and postprocessing recovery); these are detailed in Section~\ref{sec:ablation}.

\subsection{Main Results}
\label{sec:results}

\begin{table*}[t]
\centering
\caption{Comparison of answer quality, token usage, and latency across benchmarks}
\label{tab:gpt_hyve_full_verified}
\scriptsize
\setlength{\tabcolsep}{2.5pt}
\resizebox{\textwidth}{!}{%
\begin{tabular}{llccccccccccc}
\toprule
& \textbf{Model} & \textbf{Cert-QA}$^G$ & \textbf{Runbook}$^G$ & \textbf{Line}$^S$ & \textbf{Bar}$^S$ & \textbf{Anom}$^R$ & \textbf{Sum}$^G$ & \textbf{Canvas}$^R$ & \textbf{RB-Text}$^G$ & \textbf{RB-JSON}$^R$ & \textbf{Hard}$^R$ & \textbf{TOON-QA}$^E$ \\
\midrule
\multirow{4}{*}{\rotatebox{90}{Score$\uparrow$}}
& GPT-5 & \textbf{4.49} & 4.55 & 0.68 & 0.85 & 3.28 & 3.82 & 4.96 & 4.67 & 4.84 & 4.33 & 0.96 \\
& GPT-5 + HYVE & \textbf{4.49} & \textbf{4.58} & \textbf{0.97} & \textbf{1.00} & \textbf{3.77} & \textbf{3.85} & \textbf{4.96} & \textbf{4.77} & \textbf{4.92} & \textbf{5.00} & \textbf{0.98} \\
\cmidrule(lr){2-13}
& GPT-4.1 & 4.36 & 4.26 & 0.41 & \textbf{0.99} & 3.22 & 4.43 & 4.94 & 4.88 & 4.78 & 4.04 & 0.47 \\
& GPT-4.1 + HYVE & \textbf{4.38} & \textbf{4.32} & \textbf{0.95} & \textbf{0.99} & \textbf{4.03} & \textbf{4.46} & \textbf{4.95} & \textbf{4.89} & \textbf{4.92} & \textbf{5.00} & \textbf{0.93} \\
\midrule
\multirow{4}{*}{\rotatebox{90}{Token$\downarrow$}}
& GPT-5 & \textbf{3.1M} & 1.1M & 4.2M & 1.8M & 18.5M & 381.3K & 122.8M & 71.7K & 142.8K & 80.5K & 1.4M \\
& GPT-5 + HYVE & 3.1M & \textbf{1.1M} & \textbf{748.0K} & \textbf{251.8K} & \textbf{10.9M} & \textbf{372.4K} & \textbf{38.2M} & \textbf{56.8K} & \textbf{132.7K} & \textbf{20.3K} & \textbf{182.4K} \\
\cmidrule(lr){2-13}
& GPT-4.1 & \textbf{1.1M} & 289.7K & 3.0M & 1.5M & 15.8M & 209.2K & 123.4M & 61.5K & 116.1K & 75.1K & 1.3M \\
& GPT-4.1 + HYVE & 1.1M & \textbf{286.9K} & \textbf{369.4K} & \textbf{132.5K} & \textbf{8.9M} & \textbf{206.0K} & \textbf{35.1M} & \textbf{48.4K} & \textbf{107.3K} & \textbf{15.1K} & \textbf{153.8K} \\
\midrule
\multirow{4}{*}{\rotatebox{90}{Lat.(s)$\downarrow$}}
& GPT-5 & 8.10 & 38.89 & 125.11 & 75.27 & 28.73 & 23.66 & 10.45 & 20.60 & 7.39 & 19.45 & 5.48 \\
& GPT-5 + HYVE & \textbf{7.34} & \textbf{36.79} & \textbf{39.79} & \textbf{12.73} & \textbf{20.71} & \textbf{22.23} & \textbf{8.99} & \textbf{12.73} & \textbf{7.19} & \textbf{16.96} & \textbf{2.71} \\
\cmidrule(lr){2-13}
& GPT-4.1 & 3.21 & 7.58 & 47.86 & 32.91 & 7.49 & 5.57 & 3.22 & \textbf{2.74} & 2.40 & 8.07 & \textbf{1.35} \\
& GPT-4.1 + HYVE & \textbf{2.91} & \textbf{7.18} & \textbf{8.93} & \textbf{3.11} & \textbf{4.38} & \textbf{4.99} & \textbf{3.00} & 4.53 & \textbf{2.11} & \textbf{5.43} & 2.33 \\
\bottomrule
\end{tabular}
}
\par\addvspace{3mm}
\parbox{\textwidth}{\scriptsize
\textbf{GPT-5 / GPT-4.1}: Baseline models with standard JSON serialization; structured data passed directly to the LLM without optimization.
\textbf{+ HYVE}: Models augmented with the full HYVE pipeline (hybrid-view transformation, reference-guided truncation,
  queryable data backfilling and bounded SQL-augmented reasoning). \\[1pt]
$^S$Similarity (max 1.0); $^E$ExactMatch (max 1.0); $^G$GenericJudge, $^R$ReasoningJudge (max 5.0). Token values in K (thousands) or M (millions) with 1 decimal precision.}
\end{table*}

Table~\ref{tab:gpt_hyve_full_verified} summarizes our main results across 11 benchmarks.
Overall, HYVE reduces token consumption by 50--90\% on datasets with large structured inputs, while maintaining or improving answer quality across all tasks.
Latency reductions generally follow, with two exceptions where Mode~3's additional LLM call increases latency: TOON-QA and RB-Text (discussed below).
We organize our analysis by task category below.

\paragraph{Knowledge QA and Text Summarization}
The Cert-QA, Runbook, Sum, and RB-Text datasets evaluate domain knowledge and text-generation capabilities, where inputs contain moderate context and outputs are free-form text.
On these tasks, HYVE achieves comparable quality to the baselines: for GPT-5, Cert-QA remains at 4.49 and Runbook improves slightly from 4.55 to 4.58; for GPT-4.1, Cert-QA changes from 4.36 to 4.38 and Runbook from 4.26 to 4.32.
These differences are small and well within the range of normal LLM output variance.
Token usage and latency show minimal change on Cert-QA, Runbook, and Sum, as these datasets do not contain the long arrays that benefit from hybrid-view compression.
RB-Text is the exception: while quality remains comparable, GPT-4.1 latency increases from 2.74s to 4.53s due to Mode~3 overhead (discussed below).
These results confirm that HYVE introduces no regression on standard QA tasks, validating the safety of our preprocessing and postprocessing pipeline.

\begin{figure}[t]
\centering
\includegraphics[width=0.98\columnwidth]{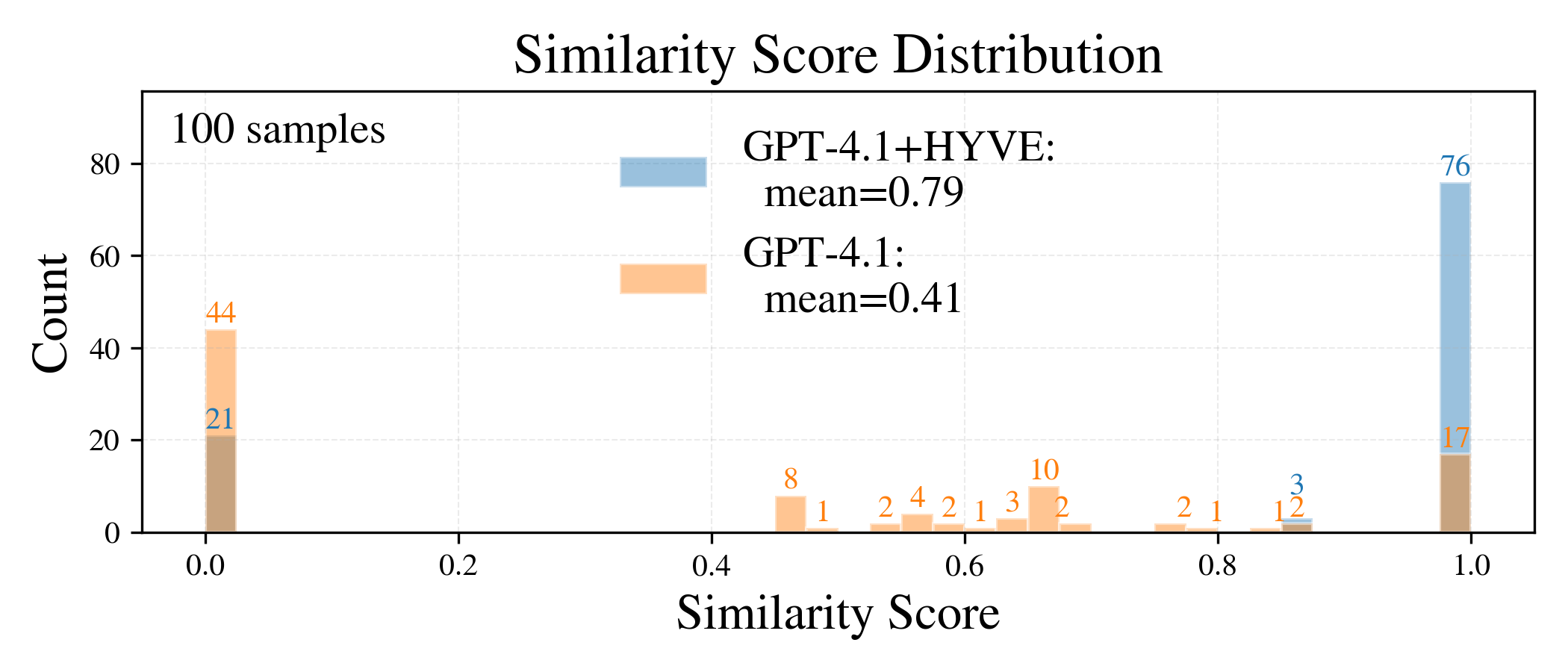}\par
\vspace{2mm}
\includegraphics[width=0.98\columnwidth]{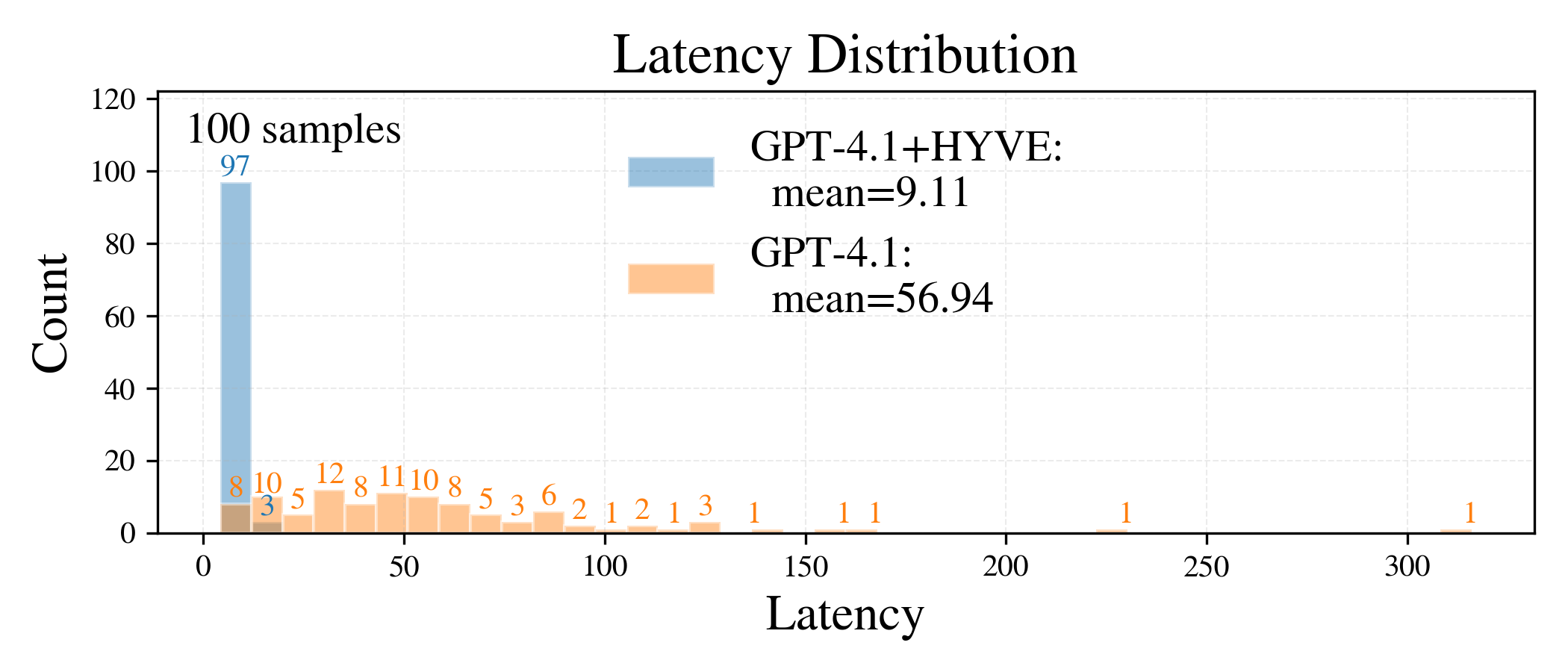}\par
\caption{Per-sample distributions on the Line chart dataset ($n=100$). HYVE raises 76\% of samples to perfect similarity (1.0), compared with 17\% for
  the baseline, while reducing median latency from 47.9s to 8.9s.}
  \label{fig:line-chart-hist}
\end{figure}

\paragraph{Structured Chart Generation}
The Line and Bar datasets require generating JSON chart specifications containing long data-point arrays.
Here HYVE yields substantial improvements across all metrics.
For GPT-5, Line chart similarity increases from 0.68 to 0.97 (+43\%), and Bar chart similarity from 0.85 to 1.00 (+18\%).
Token usage drops dramatically: Line from 4.2M to 0.75M tokens (--82\%), Bar from 1.8M to 0.25M tokens (--86\%).
Latency decreases as well: Line from 125s to 40s (--68\%), and Bar from 75s to 13s (--83\%).
The quality gains come from data backfilling, which restores arrays that would otherwise be truncated during generation.
Figure~\ref{fig:line-chart-hist} illustrates the per-sample distributions on the Line dataset: HYVE 
concentrates 76\% of samples at perfect similarity (1.0), whereas the baseline shows a bimodal distribution 
with 44\% of samples failing entirely (similarity 0.0). Latency distributions exhibit similar concentration, 
with HYVE achieving consistently low response times (median 8.9s) compared to the baseline's long-tailed distribution (median 47.9s).

\begin{figure}[t]
\centering
\includegraphics[width=0.98\columnwidth]{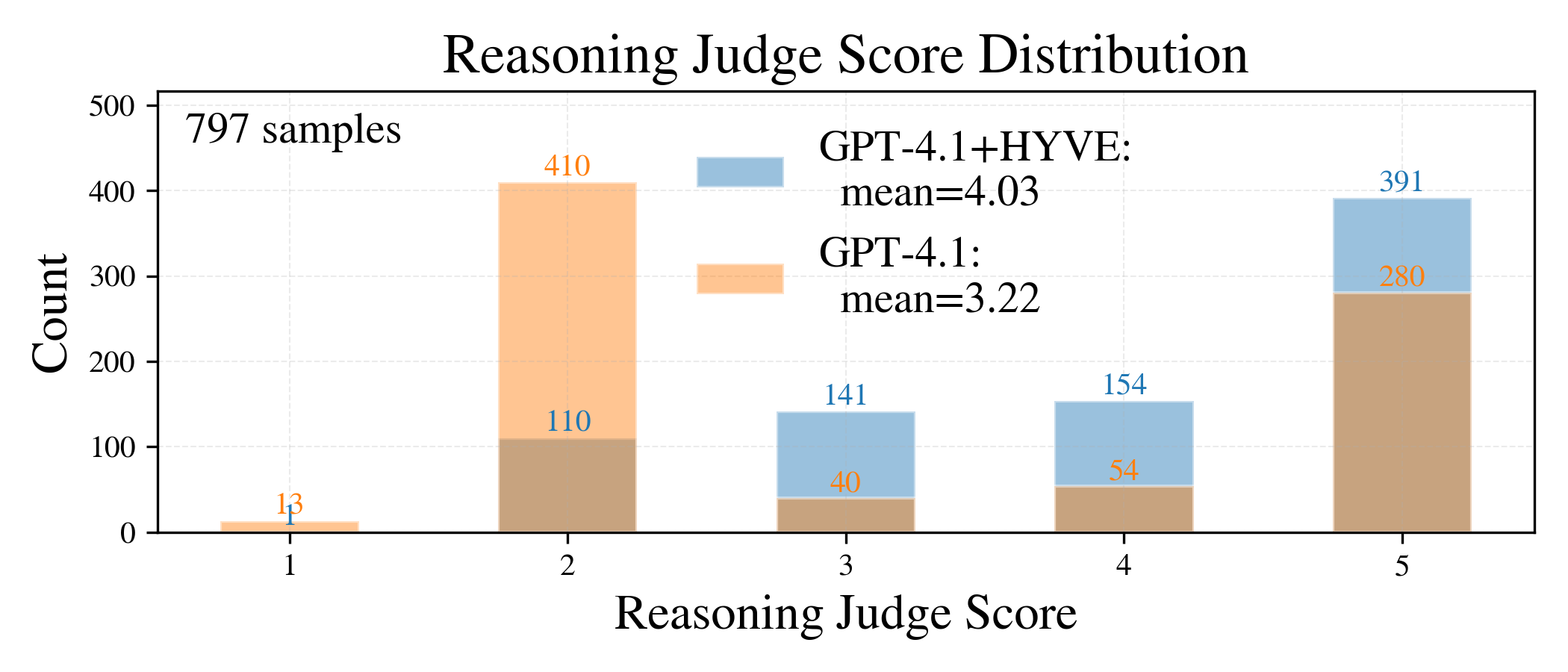}\par
\vspace{2mm}
\includegraphics[width=0.98\columnwidth]{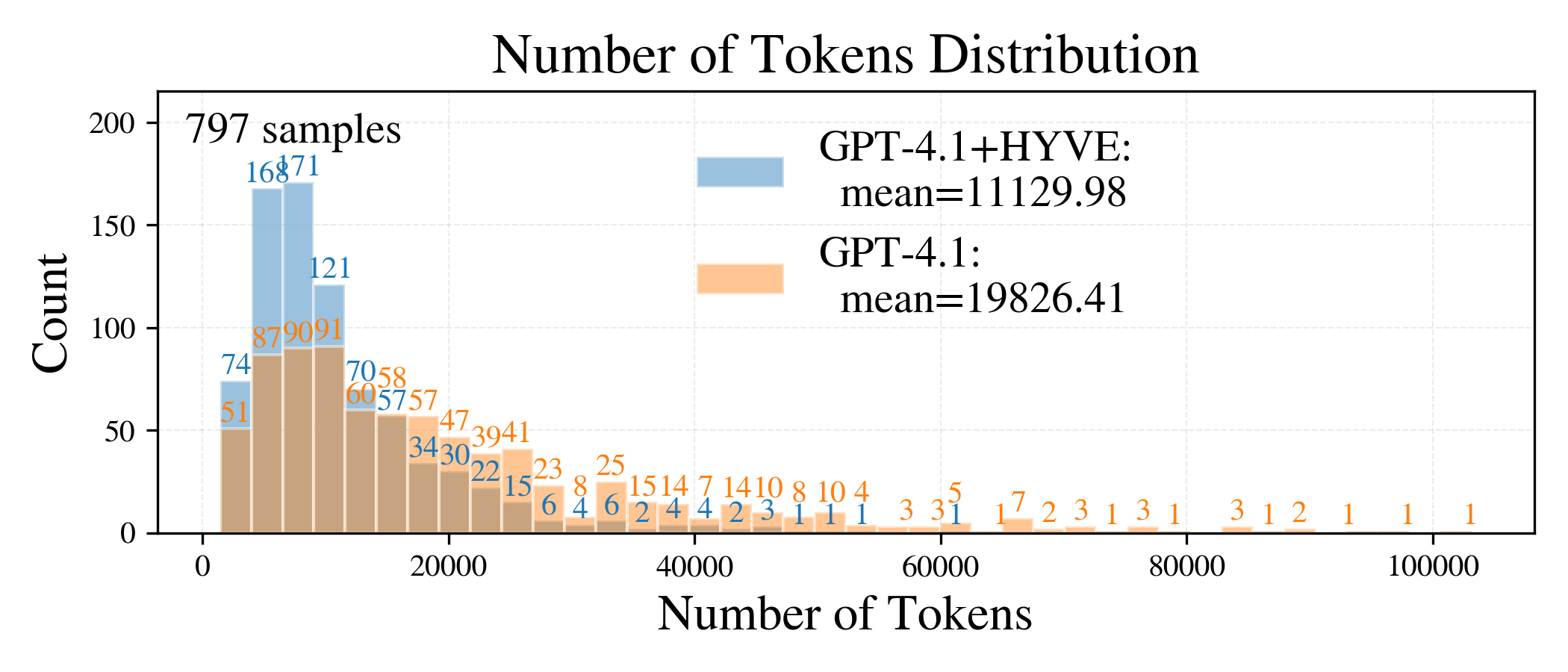}\par
\caption{Per-sample distributions on the Anom dataset (n=797). HYVE improves reasoning quality while reducing token consumption through the anomaly operator's high-signal summarization.}
\label{fig:anom-hist}
\end{figure}

\paragraph{Reasoning over Large Structured Context}
The Anom, Canvas, RB-JSON, and Hard datasets involve deeply nested JSON inputs and require structured reasoning outputs.
For GPT-5, Anom increases from 3.28 to 3.77, benefiting from the anomaly operator that extracts high-signal summaries from raw time-series data, and Hard improves from 4.33 to 5.00, attributing to the relational structure preserved in the row view for multi-hop reasoning.
On Canvas and RB-JSON, quality remains broadly stable: Canvas stays in the 4.94--4.96 range, and RB-JSON improves modestly from 4.84 to 4.92.
Token savings are substantial: Canvas drops from 122.8M to 38.2M tokens (--69\%), Anom from 18.5M to 10.9M (--41\%), and Hard from 80.5K to 20.3K (--75\%).
These results demonstrate that hybrid-view compression effectively reduces large nested contexts without sacrificing the information needed for accurate reasoning.
Figure~\ref{fig:anom-hist} shows per-sample distributions on the Anom dataset: HYVE shifts reasoning scores toward higher values while using significantly fewer tokens per sample.

\paragraph{Range-Query Reasoning with Exact-Match Evaluation}
While the preceding benchmarks mainly test point-query behavior, where retrieving a single record or field is often sufficient, TOON-QA~\cite{toon} targets range-query reasoning and is evaluated with a deterministic exact-match metric that requires no LLM judge. Of its 154 questions, 75\% require scanning, filtering, or aggregating over the \emph{entire} structured payload, while the remaining 25\% serve as field-retrieval baselines.

TOON-QA highlights two complementary benefits of HYVE. For GPT-4.1, HYVE raises exact-match accuracy from 0.47 to 0.93 (+98\%) while reducing token usage from 1.3M to 153.8K ($-$88\%). For GPT-5, the quality gain is smaller, from 0.96 to 0.98, because the baseline is already near ceiling. Even so, HYVE still substantially improves efficiency: token usage drops from 1.4M to 182.4K ($-$87\%), and latency falls from 5.48s to 2.71s ($-$51\%).

These results show that HYVE helps in two regimes: it can dramatically improve answer accuracy for weaker models on range-query reasoning, and it can substantially reduce latency and token cost even when stronger models already achieve high baseline accuracy. Notably, GPT-4.1 + HYVE (0.93) approaches GPT-5 without HYVE (0.96), suggesting that structured context engineering can partially offset model-capability differences on aggregation-style tasks.

The ablation study (Table~\ref{tab:hyve_ablation}) confirms that SQL-augmented reasoning (Mode~3) is the decisive component for TOON-QA: disabling all postprocessing (Full $-$ post) drops accuracy to 0.38 ($-$59\%). In contrast, preprocessing ablations have milder effects: removing ranking yields 0.91 ($-$2\%) and disabling truncation yields 0.87 ($-$6\%). This pattern suggests that the model can still perform aggregation with imperfect context, but the additional LLM call over queried evidence is essential for closing the remaining accuracy gap to 0.93.

TOON-QA also illustrates the latency trade-off of Mode~3. For GPT-5, HYVE reduces latency from 5.48s to 2.71s ($-$51\%) because token savings outweigh the cost of the extra LLM call. For GPT-4.1, however, latency rises from 1.35s to 2.33s because the baseline is already fast and the second call dominates. A similar trade-off appears on RB-Text, where HYVE's latency (4.53s) exceeds the baseline (2.74s) for the same reason; on all remaining benchmarks, HYVE consistently reduces latency.

\subsection{Ablation Study}
\label{sec:ablation}

We conduct ablation studies to isolate the contribution of each HYVE component.
Table~\ref{tab:hyve_ablation} organizes ablations along two dimensions.
\emph{Preprocessing ablation}: we selectively disable ranking (Full $-$ rank), truncation (Full $-$ trunc), or swap the serialization format (Full $+$ TOON), while keeping the full postprocessing pipeline (Mode~1+2+3) active.
\emph{Postprocessing ablation}: we disable all postprocessing (Full $-$ post).
We evaluate on tasks where these components have measurable impact: chart generation (Bar), anomaly detection (Anom), slot-filling (Canvas), text summarization (RB-Text), multi-hop reasoning (Hard), and range-query reasoning (TOON-QA).

\begin{table}[t]
\centering
\caption{Ablation study of HYVE with GPT-4.1}
\label{tab:hyve_ablation}
\scriptsize
\setlength{\tabcolsep}{3pt}
\resizebox{\columnwidth}{!}{%
\begin{tabular}{llcccccc}
\toprule
& \textbf{Variant} & \textbf{Bar}$^S$ & \textbf{Anom}$^R$ & \textbf{Canvas}$^R$ & \textbf{RB-Text}$^G$ & \textbf{Hard}$^R$ & \textbf{TOON-QA}$^E$ \\
\midrule
\multirow{5}{*}{\rotatebox{90}{Score$\uparrow$}}
& Full  & \textbf{0.99} & \textbf{4.03} & \textbf{4.95} & \textbf{4.89} & \textbf{5.00} & \textbf{0.93} \\
& Full $-$ rank  & \textbf{0.99} & \textbf{4.03} & \textbf{4.95} & 4.56 & \textbf{5.00} & 0.91 \\
& Full $-$ trunc  & 0.73 & 3.83 & 4.93 & 4.56 & 4.10 & 0.87 \\
& Full $+$ TOON  & 0.97 & 4.02 & 4.85 & 4.67 & \textbf{5.00} & 0.92 \\
& Full $-$ post  & 0.06 & 3.41 & 4.94 & 4.78 & 3.93 & 0.38 \\
\midrule
\multirow{5}{*}{\rotatebox{90}{Token$\downarrow$}}
& Full  & 132.5K & 8.9M & 35.1M & 48.4K & 15.1K & 153.8K \\
& Full $-$ rank  & 115.1K & 8.9M & 37.8M & 48.5K & 16.2K & 255.8K \\
& Full $-$ trunc  & 928.7K & 15.7M & 118.8M & 57.8K & 74.7K & 361.2K \\
& Full $+$ TOON  & \textbf{91.7K} & 8.9M & \textbf{28.9M} & 45.5K & \textbf{12.8K} & 143.2K \\
& Full $-$ post  & 132.3K & \textbf{6.2M} & 34.2M & \textbf{33.7K} & 14.8K & \textbf{94.5K} \\
\midrule
\multirow{5}{*}{\rotatebox{90}{Lat.(s)$\downarrow$}}
& Full  & 3.11 & 4.38 &  3.00 & 4.53 & 5.43 & 2.33 \\
& Full $-$ rank  & \textbf{2.65} & 4.91 & 4.12 & \textbf{2.39} & 8.54 & 2.56 \\
& Full $-$ trunc  & 24.48 & 5.73 & 3.33 & 3.31 & 10.80 & 1.91 \\
& Full $+$ TOON  & 3.94 & 4.76 & 3.03 & 3.27 & 5.39 & 1.86 \\
& Full $-$ post  & 3.14 & \textbf{3.20} & \textbf{2.98} & 2.68 & \textbf{5.25} & \textbf{0.97} \\
\bottomrule
\end{tabular}
}
\par\addvspace{3mm}
\parbox{\columnwidth}{\scriptsize\raggedright
Each variant changes one component; all others remain at their defaults. \\[1pt]
\textbf{Full}: Complete HYVE pipeline (Mode~1+2+3). \\[1pt]
\textbf{Full $-$ rank}: Ranking disabled; postprocessing SQL queries remain enabled; truncation keeps only the first $k$ elements (prefix-only).\\
\textbf{Full $-$ trunc}: Truncation disabled; full-length arrays passed to the LLM.\\
\textbf{Full $+$ TOON}: TOON encoding~\cite{toon} instead of beautified JSON. \\[1pt]
\textbf{Full $-$ post}: Postprocessing disabled; preprocessing, including ranking, remains enabled; output is taken directly from the LLM.\\[1pt]
$^S$Similarity (max 1.0); $^E$ExactMatch (max 1.0); $^G$GenericJudge, $^R$ReasoningJudge (max 5.0). Tokens in K/M.\par}
\end{table}

\paragraph{Ranking}
  Disabling reference-guided ranking reduces quality on tasks that require selective attention to relevant data. Without ranking, query-relevant records
  are less likely to appear in the visible context, so the system must rely more heavily on SQL-based recovery when the needed evidence lies beyond the
  retained subset. RB-Text drops from 4.89 to 4.56 (--7\%) and TOON-QA from 0.93 to 0.91 (--2\%), indicating that exposing the most relevant records
  directly in context is slightly more robust than recovering them indirectly through additional SQL queries.

  Token usage changes less consistently. It decreases slightly on Bar (132.5K$\to$115.1K) but increases on Canvas (35.1M$\to$37.8M), Hard (15.1K$
  \to$16.2K), and especially TOON-QA (153.8K$\to$255.8K), likely because more cases require SQL-based recovery once relevant records are no longer surfaced
  in the prompt.


\paragraph{Truncation}
Without truncation, token usage increases dramatically: 
Canvas rises from 35.1M to 118.8M (+239\%), Bar from 132.5K to 928.7K (+601\%), Hard from 15.1K to 74.7K (+395\%), and Anom from 8.9M to 15.7M (+76\%). Latency increases
  accordingly.
  Disabling truncation also degrades quality on Bar (0.73 vs.\ 0.99) and Hard (4.10 vs.\ 5.00, --18\%),
  suggesting that overly long contexts may induce context rot, causing the model to
  miss relevant structure. On Canvas, by contrast, the no-truncation variant nearly matches HYVE’s quality (4.93 vs.\ 4.95) despite substantial token increases, likely because
  Canvas inputs often contain relatively clear answers.
  These results confirm that truncation primarily serves token efficiency; the large overhead
  on data-intensive tasks (3--7$\times$ increases) underscores the
  cost of removing it in long-context settings.

\paragraph{Encoding}
Replacing beautified JSON with TOON encoding yields token savings: 
Canvas (-18\%, 35M$\to$29M), RB-Text (-6\%, 48K$\to$46K), and Hard (-15\%, 15K$\to$13K). 
However, TOON can degrade output quality. We attribute this to training-data bias: current LLMs are 
  predominantly trained on JSON-rich corpora,
  such as code repositories, API responses, and configuration files.
  HYVE therefore uses beautified JSON by default.

\paragraph{Postprocessing}
Disabling all postprocessing (Full $-$ post, i.e., Mode~1 only) reveals the combined impact of Mode~2 and Mode~3.
Chart generation degrades severely: Bar chart similarity drops from 0.99 to 0.06 (--94\%), confirming that LLMs often preserve the overall output schema while truncating data arrays mid-generation; without Mode~2's template-based backfilling, the missing values cannot be recovered.
Tasks relying on SQL-augmented reasoning (Mode~3) also degrade substantially: Anom drops from 4.03 to 3.41 (--15\%), as the anomaly-detection operator (exposed as a \texttt{DETECT\_ANOMALY} SQL function; see Section~\ref{sec:anomaly}) is no longer available.
Hard drops from 5.00 to 3.93 (--21\%), confirming that multi-hop reasoning benefits from the SQL query interface.
TOON-QA, which requires range-query reasoning over the datastore, falls from 0.93 to 0.38 (--59\%).
Because Anom triggers Mode~3 on nearly every sample, disabling postprocessing also reduces token usage from 8.9M to 6.2M (--30\%) and latency from 4.38s to 3.20s (--27\%), reflecting the overhead of the additional LLM call.


  

\subsection{Aligning Context Engineering and Domain Adaptation}
  \label{sec:dnm}

  The preceding experiments pair HYVE with general-purpose LLMs. A complementary direction is to combine HYVE with domain-adaptive post-training.

  DNM (Deep Network Model) is a proprietary LLM developed at Cisco on networking data. Even without HYVE, it outperforms frontier general-purpose models
  on representative networking benchmarks, with especially strong results on the CCIE (Cisco Certified Internetwork Expert) dataset (91\% vs.\ 88\% for
  GPT-5)~\cite{dnm}. This advantage is consistent with its domain-specialized training corpus.
  
 Adding HYVE on top of DNM yields further gains, especially on analytics-intensive benchmarks. On Runbook, DNM + HYVE reaches 4.61, compared with 4.58
  for GPT-5 + HYVE and 4.32 for GPT-4.1 + HYVE. These results suggest that context engineering at inference time and domain-adaptive training at training
  time are complementary. HYVE restructures the input representation, while post-training equips the model with priors that better exploit that
  representation. This alignment could be strengthened further by incorporating HYVE-style intermediate representations directly into the post-training
  data.

\section{Related Work and Limitations}
\label{sec:related}
\subsection{Related Work}
We situate HYVE at the intersection of structured-data reasoning, prompt construction, and database-inspired data organization. The most relevant prior work falls into five threads.

\noindent\textit{LLMs for Structured Data Reasoning:}
Large language models have been applied to a variety of structured-data tasks. Table question-answering systems such as TAPAS~\cite{tapas} and TaBERT~\cite{table-serialize} pre-train on table-text pairs to learn joint representations. Text-to-SQL benchmarks such as Spider~\cite{spider} and BIRD~\cite{bird} evaluate cross-domain semantic parsing, while recent prompting strategies~\cite{text2sql} improve LLM performance on SQL generation. DATER~\cite{dater} uses LLMs to decompose large tables and complex questions for more tractable reasoning. These works primarily study \emph{reasoning accuracy} over already structured inputs, typically tables or databases. HYVE addresses a complementary problem: how to transform raw prompt strings containing nested JSON/AST payloads into a representation that preserves completeness under token budgets, even when no pre-existing database is available.

\noindent\textit{Structured Data Serialization:}
The way structured data is serialized into text can strongly affect LLM performance. Sui et al.~\cite{sui2024table} systematically benchmark serialization formats such as Markdown, HTML, JSON, and CSV, showing that format choice affects both accuracy and token efficiency. TaBERT~\cite{table-serialize} linearizes tables row by row with special tokens to preserve structure. These approaches treat serialization largely as a fixed preprocessing step. By contrast, HYVE introduces a \emph{hybrid view} that combines column-oriented organization for analytics tasks with row-oriented sampling for retrieval tasks, enabling task-adaptive truncation rather than static formatting.

\noindent\textit{Prompt Compression and Context Reduction:}
  Several techniques reduce prompt length to fit within context windows. LLMLingua~\cite{llmlingua} and LongLLMLingua~\cite{longllmlingua} use smaller
  language models to identify and prune less informative tokens, while Selective Context~\cite{selective-context} applies entropy-based filtering to remove
  redundant content. These methods rely on semantic modeling to decide what to discard and therefore perform \emph{lossy compression}: once content is
  removed, it cannot be recovered. HYVE instead adopts \emph{recoverable compression} without LLM-based semantic compression, preserving a trusted visible
  subset while keeping the omitted portion accessible through the datastore for deterministic recovery when needed.

    \noindent\textit{Compression-Based Agent Memory:}
  Recent coding agents and agent runtimes, including OpenClaw~\cite{openclaw_agent}, Claude Code~\cite{claude_code}, Codex~\cite{codex_cli}, Gemini
  CLI~\cite{gemini_cli}, OpenCode~\cite{opencode_agent}, and the Pi project~\cite{pi_coding_agent}, manage growing interaction histories primarily
  through compaction by summarization. Older turns, tool traces, and intermediate results are compressed into shorter summaries so
  subsequent turns fit within the model's context window. This approach is pragmatic and widely adopted. HYVE differs in both setting and mechanism: rather than summarizing machine-generated payloads, it reorganizes them into hybrid views and preserves omitted content in a datastore for exact recovery.

\noindent\textit{Constrained Generation and Output Repair:}
Ensuring well-formed structured outputs from LLMs has received significant attention. Constrained decoding methods such as PICARD~\cite{constrained-decoding} enforce grammar constraints during autoregressive generation. LMQL~\cite{lmql} provides a query language for declarative prompting with type constraints, and grammar-constrained decoding~\cite{grammar-constrained} uses formal grammars to guide generation toward valid structured outputs. Synchromesh~\cite{json-repair} focuses on repairing syntactic errors in generated code or JSON-like outputs. These approaches emphasize \emph{syntactic correctness} or \emph{schema conformance}. HYVE addresses a different failure mode: outputs that are
  structurally valid yet \emph{data-incomplete} because arrays have been truncated. 

\noindent\textit{Hybrid Storage in Database Systems:}
Column-oriented storage~\cite{cstore} organizes data by attribute rather than by record, enabling efficient analytical queries over large datasets. Hybrid transactional/analytical processing (HTAP) systems~\cite{zhang2024htap,huang2020tidb,yang2020f1} combine row stores for transactional workloads with column stores for analytical workloads, thereby avoiding costly ETL pipelines. HYVE draws inspiration from this principle of workload-adaptive data organization: its hybrid view uses a column-oriented representation for analytics-oriented tasks such as anomaly detection and statistical summarization, and a row-oriented representation for retrieval-oriented tasks such as selecting relevant entities and relationships. This parallel suggests that workload-aware data organization remains valuable even when the execution engine is an LLM rather than a database system.

\subsection{Limitations}
  \label{sec:limitations}

  \noindent\textit{Short-Term vs. Long-Term Memory:}
  HYVE is deliberately request-scoped and therefore provides only \emph{short-term memory}: its datastore, hybrid views, and postprocessing state exist
  only within a single request. By contrast, long-term memory systems such as LongMem~\cite{longmem}, MemGPT~\cite{memgpt}, and Mem0~\cite{mem0} maintain or retrieve information
  across longer horizons using external memory stores, controllers, or retrieval policies. These approaches are orthogonal to HYVE: HYVE focuses on
  faithful within-request organization, truncation, and recovery of machine data, while long-term memory focuses on persistence across sessions. A
  natural future direction is to combine the two.

  \noindent\textit{Beyond JSON Objects:}
  HYVE assumes that repetitive input structures are already represented as JSON objects or Python/AST literals. However, some system logs contain
  structured components embedded in free-form text. A useful extension would be to automatically convert such components into JSON so HYVE can process
  them more effectively.

\section{Conclusion}
\label{sec:conclusion}
We presented \textsc{HYVE} (HYbrid ViEw), a framework for analytics-oriented context engineering over machine-data-heavy prompts. HYVE addresses a central mismatch between raw machine data and LLM reasoning by reorganizing nested payloads into hybrid row and column views, retaining full-fidelity data in a request-scoped datastore, and recovering omitted information through queryable data backfilling or bounded SQL-augmented reasoning.

The key insight is that context reduction for machine data should be \emph{structural} and \emph{recoverable}, rather than semantic and lossy. HYVE preserves data fidelity while substantially reducing visible prompt size, enabling the LLM to reason over compact, high-signal representations without losing access to the complete underlying evidence.

Across real-world networking workloads, HYVE reduces token usage by 50--90\% while preserving 
or improving answer quality. 
HYVE shows that database-inspired organization and delayed querying provide a practical foundation for scalable LLM analytics over machine data.

\bibliographystyle{\bibstylename}
\bibliography{ace-references}

@misc{toon,
  author = {{TOON Format Contributors}},
  title = {{TOON}: Token-Oriented Object Notation},
  year = {2025},
  howpublished = {\url{https://github.com/toon-format/toon}},
  note = {Includes the TOON Retrieval Accuracy Benchmark. Accessed: 2026}
}

@misc{aceblog,
  title = {{Analytics Context Engineering for LLM}},
  year = {2024},
  howpublished = {\url{https://blogs.cisco.com/ai/analytics-context-engineering-for-llm}},
  note = {February 3, 2026}
}

@misc{dnm,
  title = {{Cisco Deep Network Model: Purpose built intelligence for networking}},
  year = {2026},
  howpublished = {\url{https://blogs.cisco.com/ai/cisco-deep-network-model-overview}},
  note = {February 5, 2026}
}

@misc{vercel,
  author = {{Vercel}},
  title = {Building Filesystem Agents},
  howpublished = {\url{https://vercel.com/academy/filesystem-agents}},
  year = {2025}
}

@misc{claude,
  author       = {{Anthropic}},
  title        = {The {Claude 3} Model Family: {Opus}, {Sonnet}, {Haiku}},
  year         = {2024},
  howpublished = {\url{https://www-cdn.anthropic.com/de8ba9b01c9ab7cbabf5c33b80b7bbc618857627/Model_Card_Claude_3.pdf}}
}

@misc{jsonpath,
  title = {{JSONPath: Query Expressions for JSON}},
  howpublished = {\url{https://www.rfc-editor.org/rfc/rfc9535}},
}

@misc{openai_responses_api,
    title        = {{OpenAI API Reference: Responses}},
    howpublished = {\url{https://platform.openai.com/docs/api-reference/responses}},
    note         = {Accessed: 2026-03-25},
    year         = {2026}
  }

@misc{openai_chat_completions_api,
    title        = {{OpenAI API Reference: Chat Completions}},
    howpublished = {\url{https://platform.openai.com/docs/api-reference/chat/create-chat-completion}},
    note         = {Accessed: 2026-03-25},
    year         = {2026}
  }

@misc{anthropic_messages_api,
    author       = {{Anthropic}},
    title        = {Anthropic API: Messages Examples},
    howpublished = {\url{https://docs.anthropic.com/en/api/messages-examples}},
    note         = {Accessed: 2026-03-25},
    year         = {2026}
  }

@misc{anthropic_openai_sdk_compatibility,
    author       = {{Anthropic}},
    title        = {OpenAI SDK Compatibility},
    howpublished = {\url{https://docs.anthropic.com/en/api/openai-sdk}},
    note         = {Accessed: 2026-03-25},
    year         = {2026}
  }

@misc{openclaw_agent,
    author       = {{OpenClaw}},
    title        = {OpenClaw},
    howpublished = {\url{https://openclaw.ai/}},
    note         = {Official website. Accessed: 2026-03-30},
    year         = {2026}
  }

@misc{claude_code,
    author       = {{Anthropic}},
    title        = {Claude Code Overview},
    howpublished = {\url{https://docs.anthropic.com/en/docs/claude-code/overview}},
    note         = {Official documentation. Accessed: 2026-03-30},
    year         = {2026}
  }

@misc{codex_cli,
    author       = {{OpenAI}},
    title        = {{OpenAI Codex CLI -- Getting Started}},
    howpublished = {\url{https://help.openai.com/en/articles/11096431}},
    note         = {Official help documentation. Accessed: 2026-03-30},
    year         = {2026}
  }

@misc{gemini_cli,
    author       = {{Google}},
    title        = {{Gemini CLI}},
    howpublished = {\url{https://github.com/google-gemini/gemini-cli}},
    note         = {Official repository. Accessed: 2026-03-30},
    year         = {2026}
  }

@misc{opencode_agent,
    author       = {{OpenCode}},
    title        = {{OpenCode}},
    howpublished = {\url{https://opencode.ai/}},
    note         = {Official website. Accessed: 2026-03-30},
    year         = {2026}
  }

@misc{pi_coding_agent,
    author       = {{Pi}},
    title        = {pi.dev},
    howpublished = {\url{https://buildwithpi.com/}},
    note         = {Official website for the Pi coding agent. Accessed: 2026-03-30},
    year         = {2026}
  }

@inproceedings{duckdb2019,
    author    = {Hannes M{\"u}hleisen and Mark Raasveldt},
    title     = {DuckDB: An Embeddable Analytical Database},
    booktitle = {Proceedings of the 2019 International Conference on Management of Data (SIGMOD '19)},
    year      = {2019},
    publisher = {ACM}
  }

@article{zhang2024htap,
  title={HTAP databases: A survey},
  author={Zhang, Chao and Li, Guoliang and Zhang, Jintao and Zhang, Xinning and Feng, Jianhua},
  journal={IEEE Transactions on Knowledge and Data Engineering},
  volume={36},
  number={11},
  pages={6410--6429},
  year={2024},
  publisher={IEEE}
}

@inproceedings{cstore,
author = {Stonebraker, Mike and Abadi, Daniel J. and Batkin, Adam and Chen, Xuedong and Cherniack, Mitch and Ferreira, Miguel and Lau, Edmond and Lin, Amerson and Madden, Sam and O'Neil, Elizabeth and O'Neil, Pat and Rasin, Alex and Tran, Nga and Zdonik, Stan},
title = {C-store: a column-oriented DBMS},
year = {2005},
isbn = {1595931546},
publisher = {VLDB Endowment},
booktitle = {Proceedings of the 31st International Conference on Very Large Data Bases},
pages = {553-564},
numpages = {12},
location = {Trondheim, Norway},
series = {VLDB '05}
}

@article{huang2020tidb,
  title={TiDB: a Raft-based HTAP database},
  author={Huang, Dongxu and Liu, Qi and Cui, Qiu and Fang, Zhuhe and Ma, Xiaoyu and Xu, Fei and Shen, Li and Tang, Liu and Zhou, Yuxing and Huang, Menglong and others},
  journal={Proceedings of the VLDB Endowment},
  volume={13},
  number={12},
  pages={3072--3084},
  year={2020},
  publisher={VLDB Endowment}
}

@article{yang2020f1,
  title={F1 Lightning: HTAP as a Service},
  author={Yang, Jiacheng and Rae, Ian and Xu, Jun and Shute, Jeff and Yuan, Zhan and Lau, Kelvin and Zeng, Qiang and Zhao, Xi and Ma, Jun and Chen, Ziyang and others},
  journal={Proceedings of the VLDB Endowment},
  volume={13},
  number={12},
  pages={3313--3325},
  year={2020},
  publisher={VLDB Endowment}
}

@inproceedings{table-serialize,
  author    = {Yin, Pengcheng and Neubig, Graham and Yih, Wen-tau and Riedel, Sebastian},
  title     = {{TaBERT}: Pretraining for Joint Understanding of Textual and Tabular Data},
  booktitle = {Proceedings of the 58th Annual Meeting of the Association for Computational Linguistics},
  pages     = {8413--8426},
  year      = {2020},
  doi       = {10.18653/v1/2020.acl-main.745},
}

@inproceedings{json-repair,
  author    = {Poesia, Gabriel and Polozov, Oleksandr and Le, Vu and Tiwari, Ashish and Soares, Gustavo and Meek, Christopher and Gulwani, Sumit},
  title     = {Synchromesh: Reliable Code Generation from Pre-trained Language Models},
  booktitle = {International Conference on Learning Representations},
  year      = {2022},
}

@inproceedings{constrained-decoding,
  author    = {Scholak, Torsten and Schucher, Nathan and Bahdanau, Dzmitry},
  title     = {{PICARD}: Parsing Incrementally for Constrained Auto-Regressive Decoding from Language Models},
  booktitle = {Proceedings of the 2021 Conference on Empirical Methods in Natural Language Processing},
  pages     = {9895--9901},
  year      = {2021},
  doi       = {10.18653/v1/2021.emnlp-main.779},
}

@article{lmql,
  title={Prompting is programming: A query language for large language models},
  author={Beurer-Kellner, Luca and Fischer, Marc and Vechev, Martin},
  journal={Proceedings of the ACM on Programming Languages},
  volume={7},
  number={PLDI},
  pages={1946--1969},
  year={2023},
  publisher={ACM New York, NY, USA}
}

@inproceedings{grammar-constrained,
    title = "Grammar-Constrained Decoding for Structured {NLP} Tasks without Finetuning",
    author = "Geng, Saibo  and
      Josifoski, Martin  and
      Peyrard, Maxime  and
      West, Robert",
    editor = "Bouamor, Houda  and
      Pino, Juan  and
      Bali, Kalika",
    booktitle = "Proceedings of the 2023 Conference on Empirical Methods in Natural Language Processing",
    month = dec,
    year = "2023",
    address = "Singapore",
    publisher = "Association for Computational Linguistics",
    url = "https://aclanthology.org/2023.emnlp-main.674/",
    doi = "10.18653/v1/2023.emnlp-main.674",
    pages = "10932--10952"
}

@inproceedings{llmlingua,
  author    = {Jiang, Huiqiang and Wu, Qianhui and Lin, Chin-Yew and Yang, Yuqing and Qiu, Lili},
  title     = {{LLMLingua}: Compressing Prompts for Accelerated Inference of Large Language Models},
  booktitle = {Proceedings of the 2023 Conference on Empirical Methods in Natural Language Processing},
  pages     = {13358--13376},
  year      = {2023},
  doi       = {10.18653/v1/2023.emnlp-main.825},
}

@inproceedings{longllmlingua,
    title = "{L}ong{LLML}ingua: Accelerating and Enhancing {LLM}s in Long Context Scenarios via Prompt Compression",
    author = "Jiang, Huiqiang  and
      Wu, Qianhui  and
      Luo, Xufang  and
      Li, Dongsheng  and
      Lin, Chin-Yew  and
      Yang, Yuqing  and
      Qiu, Lili",
    editor = "Ku, Lun-Wei  and
      Martins, Andre  and
      Srikumar, Vivek",
    booktitle = "Proceedings of the 62nd Annual Meeting of the Association for Computational Linguistics (Volume 1: Long Papers)",
    month = aug,
    year = "2024",
    address = "Bangkok, Thailand",
    publisher = "Association for Computational Linguistics",
    url = "https://aclanthology.org/2024.acl-long.91/",
    doi = "10.18653/v1/2024.acl-long.91",
    pages = "1658--1677"
}

@inproceedings{selective-context,
    title = "Compressing Context to Enhance Inference Efficiency of Large Language Models",
    author = "Li, Yucheng  and
      Dong, Bo  and
      Guerin, Frank  and
      Lin, Chenghua",
    editor = "Bouamor, Houda  and
      Pino, Juan  and
      Bali, Kalika",
    booktitle = "Proceedings of the 2023 Conference on Empirical Methods in Natural Language Processing",
    month = dec,
    year = "2023",
    address = "Singapore",
    publisher = "Association for Computational Linguistics",
    url = "https://aclanthology.org/2023.emnlp-main.391/",
    doi = "10.18653/v1/2023.emnlp-main.391",
    pages = "6342--6353"
}

@inproceedings{sui2024table,
  author    = {Sui, Yuan and Zhou, Mengyu and Zhou, Mingjie and Han, Shi and Zhang, Dongmei},
  title     = {Table Meets {LLM}: Can Large Language Models Understand Structured Table Data? A Benchmark and Empirical Study},
  booktitle = {Proceedings of the 17th ACM International Conference on Web Search and Data Mining},
  pages     = {645--654},
  year      = {2024},
  doi       = {10.1145/3616855.3635752},
}

@inproceedings{tapas,
    title = "{T}a{P}as: Weakly Supervised Table Parsing via Pre-training",
    author = {Herzig, Jonathan  and
      Nowak, Pawel Krzysztof  and
      M{\"u}ller, Thomas  and
      Piccinno, Francesco  and
      Eisenschlos, Julian},
    editor = "Jurafsky, Dan  and
      Chai, Joyce  and
      Schluter, Natalie  and
      Tetreault, Joel",
    booktitle = "Proceedings of the 58th Annual Meeting of the Association for Computational Linguistics",
    month = jul,
    year = "2020",
    address = "Online",
    publisher = "Association for Computational Linguistics",
    url = "https://aclanthology.org/2020.acl-main.398/",
    doi = "10.18653/v1/2020.acl-main.398",
    pages = "4320--4333"
}

@inproceedings{spider,
    title = "{S}pider: A Large-Scale Human-Labeled Dataset for Complex and Cross-Domain Semantic Parsing and Text-to-{SQL} Task",
    author = "Yu, Tao  and
      Zhang, Rui  and
      Yang, Kai  and
      Yasunaga, Michihiro  and
      Wang, Dongxu  and
      Li, Zifan  and
      Ma, James  and
      Li, Irene  and
      Yao, Qingning  and
      Roman, Shanelle  and
      Zhang, Zilin  and
      Radev, Dragomir",
    editor = "Riloff, Ellen  and
      Chiang, David  and
      Hockenmaier, Julia  and
      Tsujii, Jun{'}ichi",
    booktitle = "Proceedings of the 2018 Conference on Empirical Methods in Natural Language Processing",
    month = oct # "-" # nov,
    year = "2018",
    address = "Brussels, Belgium",
    publisher = "Association for Computational Linguistics",
    url = "https://aclanthology.org/D18-1425/",
    doi = "10.18653/v1/D18-1425",
    pages = "3911--3921"
}

@inproceedings{bird,
  author = {Li, Jinyang and Hui, Binyuan and Qu, Ge and Yang, Jiaxi and Li, Binhua and Li, Bowen and Wang, Bailin and Qin, Bowen and Geng, Ruiying and Huo, Nan and Zhou, Xuanhe and Ma, Chenhao and Li, Guoliang and Chang, Kevin C.C. and Huang, Fei and Cheng, Reynold and Li, Yongbin},
  title = {Can LLM already serve as a database interface? a big bench for large-scale database grounded text-to-SQLs},
  year = {2023},
  publisher = {Curran Associates Inc.},
  address = {Red Hook, NY, USA},
  booktitle = {Proceedings of the 37th International Conference on Neural Information Processing Systems},
  articleno = {1835},
  numpages = {28},
  location = {New Orleans, LA, USA},
  series = {NIPS '23}
}

@inproceedings{text2sql,
    title = "Enhancing Text-to-{SQL} Capabilities of Large Language Models through Tailored Promptings",
    author = "Tan, Zhao  and
      Liu, Xiping  and
      Shu, Qing  and
      Li, Xi  and
      Wan, Changxuan  and
      Liu, Dexi  and
      Wan, Qizhi  and
      Liao, Guoqiong",
    editor = "Calzolari, Nicoletta  and
      Kan, Min-Yen  and
      Hoste, Veronique  and
      Lenci, Alessandro  and
      Sakti, Sakriani  and
      Xue, Nianwen",
    booktitle = "Proceedings of the 2024 Joint International Conference on Computational Linguistics, Language Resources and Evaluation (LREC-COLING 2024)",
    month = may,
    year = "2024",
    address = "Torino, Italia",
    publisher = "ELRA and ICCL",
    url = "https://aclanthology.org/2024.lrec-main.539/",
    pages = "6091--6109"
}

@article{chandola2009anomaly,
  title={Anomaly detection: A survey},
  author={Chandola, Varun and Banerjee, Arindam and Kumar, Vipin},
  journal={ACM computing surveys (CSUR)},
  volume={41},
  number={3},
  pages={1--58},
  year={2009},
  publisher={ACM New York, NY, USA}
}

@inproceedings{dater,
author = {Ye, Yunhu and Hui, Binyuan and Yang, Min and Li, Binhua and Huang, Fei and Li, Yongbin},
title = {Large Language Models are Versatile Decomposers: Decomposing Evidence and Questions for Table-based Reasoning},
year = {2023},
isbn = {9781450394086},
publisher = {Association for Computing Machinery},
address = {New York, NY, USA},
url = {https://doi.org/10.1145/3539618.3591708},
doi = {10.1145/3539618.3591708},
booktitle = {Proceedings of the 46th International ACM SIGIR Conference on Research and Development in Information Retrieval},
pages = {174-184},
numpages = {11},
location = {Taipei, Taiwan},
series = {SIGIR '23}
}

@article{robertson2009probabilistic,
  author = {Robertson, Stephen and Zaragoza, Hugo},
  title = {The Probabilistic Relevance Framework: BM25 and Beyond},
  journal = {Foundations and Trends in Information Retrieval},
  volume = {3},
  number = {4},
  pages = {333--389},
  year = {2009},
  doi = {10.1561/1500000019},
  url = {https://doi.org/10.1561/1500000019}
}

@misc{langsmith,
  title = {{LangSmith}},
  author = {{LangChain, Inc.}},
  year = {2023},
  howpublished = {\url{https://www.langchain.com/langsmith}},
  note = {Accessed: 2026}
}

@article{longmem,
  author  = {Wang, Weizhi and Dong, Li and Cheng, Hao and Liu, Xiaodong and Yan, Xifeng and Gao, Jianfeng and Wei, Furu},
  title   = {Augmenting Language Models with Long-Term Memory},
  journal = {arXiv preprint arXiv:2306.07174},
  year    = {2023}
}

@article{memgpt,
  author  = {Packer, Charles and Fang, Vivian and Patil, Shishir G. and Lin, Kevin and Wooders, Sarah and Gonzalez, Joseph E.},
  title   = {MemGPT: Towards LLMs as Operating Systems},
  journal = {arXiv preprint arXiv:2310.08560},
  year    = {2023}
}

@article{mem0,
  author  = {Chhikara, Prateek and Khant, Dev and Aryan, Saket and Singh, Taranjeet and Yadav, Deshraj},
  title   = {Mem0: Building Production-Ready AI Agents with Scalable Long-Term Memory},
  journal = {arXiv preprint arXiv:2504.19413},
  year    = {2025}
}

\appendices

\lstdefinestyle{appendixcode}{
  basicstyle=\ttfamily\scriptsize,
  breaklines=true,
  keepspaces=true,
  columns=fullflexible,
  xleftmargin=1em,
  aboveskip=4pt,
  belowskip=4pt
}

This appendix is organized into three parts. We first present representative benchmark examples to ground the tasks summarized in the main body, then list the evaluation prompts used in our experiments, and finally provide additional implementation details on SQL prompt generation together with the execution logic of Modes~2 and~3. Unless otherwise noted, all snippets are shortened excerpts for presentation.

\section{Representative Dataset Examples}
\subsection{Cert-QA Dataset}
\label{app:cert-qa-examples}

This appendix provides representative examples from each certification level in the Cert-QA dataset.

\subsubsection{CCNA-Level Example (Entry)}

\textbf{Question:} What command is used on a Windows PC to display IP-to-MAC address mappings?

\textbf{Answer:} \texttt{arp -a}

\subsubsection{CCNP-Level Example (Advanced)}

\textbf{Question:} What is Generic Routing Encapsulation (GRE), and what was its original purpose?

\textbf{Answer:} GRE is a tunneling protocol that encapsulates packets over an IP-based network. It was originally created to provide transport for non-routable legacy protocols (like IPX) across an IP network.

\subsubsection{CCIE-Level Example (Expert, Open-Ended)}

\textbf{Question:} Cisco Jabber clients need to be able to reach several different applications to provide access to services such as voicemail, meetings, directories, and other functions. Which profile must be configured to provide these services?

\textbf{Answer:} Service profile

\subsubsection{CCIE-Level Example (Expert, Multiple-Choice)}

\textbf{Question:} What is the deployment model of the Cisco Secure Network Analytics Cognitive Analytics system?
\begin{enumerate}
\item[(a)] as a plug-in in the Cisco Secure Network Analytics Management Console
\item[(b)] in the public cloud (SaaS)
\item[(c)] on-premise as a dedicated appliance
\item[(d)] on-premise as a virtual machine
\end{enumerate}

\textbf{Answer:} (b) in the public cloud (SaaS)

\textbf{Explanation:} Cisco Secure Network Analytics Cognitive Analysis is offered as a Software as a Service product. The service is deployed in the Cisco cloud and other deployment options are not possible.

\subsubsection{Expert-Tiered Example (Advanced Topics)}

\textbf{Question:} In BGP implementations, what attribute is used to influence inbound traffic from neighbouring ASes?

\textbf{Answer:} AS-PATH prepending.

\subsection{Runbook Dataset}
\label{app:runbook-examples}

This appendix provides a representative example from the Runbook dataset.

\subsubsection{StackWise Upgrade Troubleshooting}

\textbf{Problem Description:}
\begin{enumerate}
\item Install mode in stackwise standard upgrade procedure.
\item Install mode in stackwise, if one machine (non-active switch) prompts v-mismatch. How to correctly upgrade this device?
\item If using a 4-switch stack, but these 4 switches have different versions, can all switches be upgraded directly to the specified version through the active switch?
\end{enumerate}

\textbf{Ground-Truth Runbook (excerpt):}

\begin{quote}
\textbf{StackWise Upgrade Troubleshooting}

\textit{Summary:} This playbook outlines the troubleshooting steps for upgrading a Catalyst 9300 series switch stack. The focus is on addressing potential issues encountered during the upgrade process, such as version mismatches and ensuring successful upgrades across all stack members.

\textbf{1. Initial Assessment:}
\begin{itemize}
\item Access the Active Switch CLI: Establish connectivity to the active switch within the stack. Use Telnet or SSH to access the CLI.
\item Verify Stack Status: Execute \texttt{show stack status} to confirm the operational state of the stack.
\item Check Software Versions: Run \texttt{show version} on the active switch to identify the current software version.
\end{itemize}

\textbf{2. Version Mismatch Resolution:}
\begin{itemize}
\item If a non-active switch shows v-mismatch, use \texttt{install add file <image> activate commit} from the active switch.
\item The active switch will propagate the image to mismatched members automatically.
\end{itemize}

\textbf{3. Multi-Version Stack Upgrade:}
\begin{itemize}
\item Yes, all switches can be upgraded from the active switch using install mode.
\item Execute \texttt{install add file flash:<image> activate commit} to upgrade all stack members simultaneously.
\end{itemize}
\end{quote}

\subsection{Line Chart Dataset}
\label{app:line-chart-examples}

This appendix provides a representative example from the Line chart dataset.

\subsubsection{Line Chart Example}

\textbf{Input Data (excerpt):}
\begin{lstlisting}[style=appendixcode]
[
  {"endTs": "2025-06-14T22:32:00Z", 
   "jitter": 0.23, "goodput": 100000,
   "startTs": "2025-06-14T22:31:00Z", 
   "latencyMs": 9.07, "lossPercent": 0},
  {"endTs": "2025-06-14T22:34:00Z", 
   "jitter": 0.11, "goodput": 100000,
   "startTs": "2025-06-14T22:33:00Z", 
   "latencyMs": 9.02, "lossPercent": 0},
  ...
]
\end{lstlisting}

\textbf{Expected Output (excerpt):}
\begin{lstlisting}[style=appendixcode]
{"data": {
  "name": "CDSAILineChart",
  "props": {
    "id": "meraki-loss-latency-chart",
    "axes": {"xAxis": {"type": "time"}, 
             "yAxis": {"format": {}}},
    "data": [
      {"id": "loss", "data": [
        {"x": "2025-06-14T22:31:00Z", "y": 0},
        {"x": "2025-06-14T22:33:00Z", "y": 0},
        ...]},
      {"id": "latency", "data": [
        {"x": "2025-06-14T22:31:00Z", "y": 9.07},
        ...]}
    ]}}}
\end{lstlisting}

\subsection{Bar Chart Dataset}
\label{app:bar-chart-examples}

This appendix provides a representative example from the Bar chart dataset.

\subsubsection{Bar Chart Example}

\textbf{Input Data (excerpt):}
\begin{lstlisting}[style=appendixcode]
[
  {"peer": 59, "expansionism": "sediment"},
  {"peer": 98, "expansionism": "nonsense"},
  {"peer": 81, "expansionism": "cloud"},
  ...
]
\end{lstlisting}

\textbf{Expected Output (excerpt):}
\begin{lstlisting}[style=appendixcode]
{"data": {"props": {"data": [
  {"stacks": [{"key": "peer", "value": 59}], 
   "category": "sediment"},
  {"stacks": [{"key": "peer", "value": 98}], 
   "category": "nonsense"},
  {"stacks": [{"key": "peer", "value": 81}], 
   "category": "cloud"},
  ...
]}}}
\end{lstlisting}

\subsection{Anomaly Detection Dataset}
\label{app:anom-examples}

This appendix provides a representative example from the Anom dataset.

\subsubsection{Network Path Latency Analysis}

\textbf{Task Instruction:}
\begin{quote}
Use the Network path data to identify IF there are any specific high latency nodes. Provide a summary to the user of impacted nodes.
\end{quote}

\textbf{Input Data (excerpt):}
\begin{lstlisting}[style=appendixcode]
{"test": {"testId": "281208", 
  "testName": "Synthetic Network Test", 
  "type": "network"},
 "pathVis": [{
   "agent": {"agentName": "agent-89", 
             "countryId": "JP"},
   "server": "service26.example.com:443",
   "pathTraces": [{
     "hops": [
       {"hop": 1, "ipAddress": "180.137.137.113",
        "hostname": "srv-34.fernandez.com", 
        "responseTime": 2},
       {"hop": 2, "ipAddress": "132.239.231.168",
        "hostname": "lt-24.davies.com", 
        "responseTime": 2},
       ...
     ]}]}]}
\end{lstlisting}

\textbf{Expected Output:}
\begin{lstlisting}[style=appendixcode]
{"Reasoning": [
  "Each path trace contains hops with response times.",
  "Any node with responseTime > 10ms is high latency.",
  "Iterate through each path trace and check hops.",
  "No node has responseTime exceeding 10ms threshold.",
  "Based on the data, there are no high latency nodes."
 ],
 "variables": {
  "Summary of high latency nodes": 
    "No nodes with high latency were detected."
 }}
\end{lstlisting}

\subsection{Summarization Dataset}
\label{app:sum-examples}

This appendix provides a representative example from the Sum dataset.

\subsubsection{Board Report Generation}

\textbf{Task Instruction:}
\begin{quote}
Given a JSON object that contains the current context of the `board', generate a holistic report that derives key data points, insights, timelines, incident details, resolutions, or root cause analysis. Organize your response with clearly defined sections and a table of contents.
\end{quote}

\textbf{Input Data (excerpt):}
\begin{lstlisting}[style=appendixcode]
{"board": {
  "name": "AIC 10 - Sep 27, 2025 5:42 PM",
  "metadata": {},
  "canvas": {"id": "canvas-12345", ...},
  "cards": [...],
  "conversations": [...]
}}
\end{lstlisting}

\textbf{Expected Output (excerpt):}
\begin{lstlisting}[style=appendixcode]
# Report on Board "AIC 10 - Sep 27, 2025"

## Table of Contents
1. [Introduction](#introduction)
2. [Board Overview](#board-overview)
3. [Canvas Details](#canvas-details)
4. [Cards Analysis](#cards-analysis)
5. [Conversations Historical Context]
6. [Conclusion](#conclusion)

## Introduction
This report presents an analysis of the board 
titled "AIC 10 - Sep 27, 2025 5:42 PM," 
providing insights into its components...

## Board Overview
- **Name**: AIC 10 - Sep 27, 2025 5:42 PM
- **Description**: The board is identified by 
  its timestamp, suggesting it may be part of 
  a series or larger project...
\end{lstlisting}

\subsection{Canvas Dataset}
\label{app:canvas-examples}

This appendix provides a representative example from the Canvas slot-filling dataset.

\subsubsection{Runbook Step: Network Identification}

\textbf{Task Context:}
\begin{quote}
You are an expert in networking with a CCIE certification. You are helping with running a network troubleshooting run-book. Steps involve data gathering, analyzing the data and setting variables, running commands against the network, and making decisions based on the data.
\end{quote}

\textbf{Input Data (excerpt):}
\begin{lstlisting}[style=appendixcode]
{"context": {
  "board": {
    "board_id": {"id": "cf53d6d9-ef2a-..."},
    "name": {"name": "Paul Gomez - Aug 15, 2025"},
    "metadata": {...},
    "canvas": {...},
    "cards": [...],
    "conversations": [...]
  }
},
"runbook_step": "Get the list of all networks 
  in the org. Find the network that best matches 
  'Seattle' location and return the networkId 
  and network name."
}
\end{lstlisting}

\textbf{Expected Output:}
\begin{lstlisting}[style=appendixcode]
{"core_skill_input": {
  "orgId": "928177",
  "query": "Get the list of all the networks 
    in the org. Find the network that best 
    matches 'Seattle' location and return 
    the networkId and network name.",
  "baseUrl": "https://dashboard.meraki.com/api/v1",
  "networkId": "L_4729533",
  "organizationId": "928177"
}}
\end{lstlisting}

\subsection{RB-Text Dataset}
\label{app:rb-text-examples}

This appendix provides a representative example from the RB-Text API response summarization dataset.

\subsubsection{ThousandEyes Analysis Summary}

\textbf{Task Context:}
\begin{quote}
You are an expert in networking with a CCIE certification. You are helping with running a network troubleshooting run-book. Please summarize the following response that was obtained calling an API and output it in a markdown format.
\end{quote}

\textbf{API Response (excerpt):}
\begin{lstlisting}[style=appendixcode]
{"status": "success", 
 "message": "The analysis cannot be completed 
   due to invalid start or end timestamps. 
   This typically occurs when the specified 
   time range is not available in the test 
   data or the timestamps are outside the 
   valid data retention period..."}
\end{lstlisting}

\textbf{Expected Output:}
\begin{lstlisting}[style=appendixcode]
**ThousandEyes Analysis Summary**

- Analysis could not be completed due to 
  invalid start or end timestamps.
- The specified time range is either 
  unavailable in test data or outside 
  the data retention period.
- To resolve:
  - Use the "Explain Selection" button in 
    the ThousandEyes UI for valid timestamps.
  - Ensure test ID 7802744 has data for 
    the selected period.
  - Verify timestamps are within your 
    account's data retention period.
\end{lstlisting}

\subsection{RB-JSON Dataset}
\label{app:rb-json-examples}

This appendix provides a representative example from the RB-JSON conditional expression evaluation dataset.

\subsubsection{Variable Existence Check}

\textbf{Task Context:}
\begin{quote}
You are an expert in networking with a CCIE certification. We are executing a flow chart that corresponds to a run-book used for network troubleshooting. We need to figure out the truth value of the expression in the reasoning instruction to execute the flow chart.
\end{quote}

\textbf{Reasoning Instruction:}
\begin{lstlisting}[style=appendixcode]
If "Location" is set, go to step 3.
\end{lstlisting}

\textbf{Context Variables:}
\begin{lstlisting}[style=appendixcode]
{"Application": "video server", 
 "Location": "San Francisco, CA", 
 "StartTime": "2025-11-10T00:00:00Z"}
\end{lstlisting}

\textbf{Expected Output:}
\begin{lstlisting}[style=appendixcode]
{"Reasoning": [
   "The variable 'Location' is present 
    in the provided data.",
   "The value of 'Location' is 'San Francisco, CA', 
    which means it is set (not null or empty).",
   "Therefore, the condition 'Location is set' 
    is true."
 ],
 "expression": "Location is set",
 "truth_value_of_expression": true}
\end{lstlisting}

\subsection{TOON-QA Dataset}
\label{app:toon-qa-examples}

This appendix provides representative examples from the TOON-QA dataset~\cite{toon}. Questions span field retrieval, aggregation, and multi-condition filtering over structured JSON payloads.

\subsubsection{Order Filtering and Counting}

\textbf{Input Data (excerpt):}
\begin{lstlisting}[style=appendixcode]
{"orders": [
  {"orderId": "ORD-0001",
   "customer": {"id": 1, "name": "Valerie Braun",
     "email": "name.jones@gmail.com"},
   "items": [
     {"sku": "SKU-OOH73G", "name": "Widget A",
      "quantity": 2, "price": 29.99},
     {"sku": "SKU-PLM12X", "name": "Gadget B",
      "quantity": 1, "price": 49.99},
     ...
   ],
   "status": "processing",
   "total": 109.97},
  ...
]}
\end{lstlisting}

\textbf{Question:}
\begin{quote}
How many orders have status ``processing'' and at least 3 items?
Provide only the direct answer, without any additional explanation or formatting.
\end{quote}

\textbf{Expected Answer:} \texttt{5}

\subsubsection{Employee Aggregation}

\textbf{Question:}
\begin{quote}
How many active employees have more than 5 years of experience?
Provide only the direct answer, without any additional explanation or formatting.
\end{quote}

\textbf{Expected Answer:} \texttt{63}

\subsubsection{Time-Series Lookup}

\textbf{Question:}
\begin{quote}
What was the revenue on 2025-01-04?
Provide only the direct answer, without any additional explanation or formatting.
\end{quote}

\textbf{Expected Answer:} \texttt{8357.79}

\subsection{Hard Reasoning Dataset}
\label{app:hard-examples}

This appendix provides a representative example from the Hard multi-hop reasoning dataset.

\subsubsection{Multi-Hop Test Discovery}

\textbf{Query:}
\begin{quote}
Find a ThousandEyes DNS test (type: dns-server) that is related to the Application (SharePoint) and is run from an agent in the Location (San Francisco).
\end{quote}

\textbf{Input Data (excerpt):}
\begin{lstlisting}[style=appendixcode]
{"tests": [
  {"testId": 264343, 
   "testName": "A - SharePoint - DNS - Internal",
   "type": "dns-server",
   "target": "cisco.sharepoint.com",
   "agents": [
     {"agentName": "pdx-linux-ea", "location": "Portland"},
     {"agentName": "stg-teva", "location": "San Jose"},
     {"agentName": "newark-agent", "location": "Newark, NJ"}
   ]},
  {"testId": 264347,
   "testName": "A - SharePoint - DNS - External",
   "type": "dns-server",
   "target": "cisco.sharepoint.com",
   "agents": [
     {"agentName": "phoenix-agent", "location": "Phoenix, AZ"},
     {"agentName": "sf-agent", "location": "San Francisco, CA"}
   ]},
  ...
]}
\end{lstlisting}

\textbf{Expected Output (excerpt):}
\begin{lstlisting}[style=appendixcode]
{"Reasoning": [
  "The instruction is to find a ThousandEyes DNS 
   test (type: dns-server) related to SharePoint 
   running from an agent in San Francisco.",
  "I found tests with 'SharePoint' in the name: 
   'A - SharePoint - DNS - Internal' (264343), 
   'A - SharePoint - DNS - External' (264347).",
  "'A - SharePoint - DNS - Internal' has agents in 
   Portland, San Jose, Newark - none in SF.",
  "'A - SharePoint - DNS - External' has agents in 
   Phoenix and San Francisco. This matches!",
  "The matching test is testId 264347."
 ],
 "Answer": {
   "testId": "264347",
   "testName": "A - SharePoint - DNS - External"
 }}
\end{lstlisting}

\section{Evaluation Prompts}
\label{app:eval-prompts}

This appendix provides the complete prompts used by our LLM-as-a-judge evaluators.

\subsection{GenericJudge Prompt}
\label{app:generic-judge-prompt}

The following prompt is used by the GenericJudge evaluator (GPT-4o) for open-ended quality assessment:

\begin{lstlisting}[style=appendixcode]
You are an evaluator tasked with comparing a Ground Truth Answer, 
which represents the correct solution, to a Generated Answer, 
which is the system's response to a given input question.

Follow these steps to conduct your evaluation:

1. **Read Carefully**
   Review the Question, Ground Truth Answer, and Generated Answer 
   thoroughly.

2. **Assign a Score (1-5)**
   Evaluate the Generated Answer against the Ground Truth Answer 
   using the following rubric:

   * **5 - Excellent**: Fully correct, complete, and clearly 
     articulated. Matches the Ground Truth in factual content 
     and intent with no significant errors or omissions.
   * **4 - Good**: Mostly correct and covers most key points. 
     Minor inaccuracies or omissions that don't significantly 
     affect overall quality.
   * **3 - Fair**: Partially correct. Captures some key elements 
     but misses others. May include inaccuracies or lack clarity.
   * **2 - Poor**: Largely incorrect or incomplete. Contains 
     major errors, omissions, or misinterpretations.
   * **1 - Inadequate**: Completely incorrect, irrelevant, or 
     nonsensical. Shows no meaningful understanding of the query.

3. **Output Final JSON**
   Return a valid JSON object with exactly two keys:
   * "score": an integer from 1 to 5
   * "justification": a brief explanation for the score

Here are the inputs for you to conduct your evaluation:

Question:
[BEGIN QUESTION]
{question}
[END QUESTION]

Ground Truth Answer:
[BEGIN GROUND TRUTH ANSWER]
{ground_truth}
[END GROUND TRUTH ANSWER]

Generated Answer:
[BEGIN GENERATED ANSWER]
{response}
[END GENERATED ANSWER]
\end{lstlisting}

\subsection{ReasoningJudge Prompt}
\label{app:reasoning-judge-prompt}

The following prompt is used by the ReasoningJudge evaluator (GPT-4.1) for structured JSON output assessment:

\begin{lstlisting}[style=appendixcode]
You are an evaluator tasked with assessing a System Under Test (SUT) 
JSON output against a Ground Truth JSON and a provided JSON Schema. 
Base your judgment strictly on the Ground Truth and Schema--do not 
use outside knowledge.

Follow these steps:

1. **Parse and Validate JSON**
   * Extract JSON from the SUT Output (strip code fences if present).
   * If parsing fails, assign a score of 1.
   * If parsing succeeds, validate against the Schema.

2. **Compare to Ground Truth**
   * Ignore extra fields not in the Ground Truth.
   * For each Ground Truth field, check:
     * Field exists in the SUT.
     * Type matches.
     * Value matches, using these rules:
       - Strings: exact match after trimming whitespace.
       - Numbers/booleans: exact equality.
       - Arrays (structured data): same length, element-wise 
         equality, same order.
       - Arrays (reasoning): allow grouping/splitting with 
         semantic matching; penalize missing or contradictory 
         points.
       - Objects: compare recursively.
   * For free-text fields: paraphrasing is acceptable only if 
     all factual points are preserved and no contradictions 
     are introduced.

3. **Assign a Score (1-5)**
   * **5 - Excellent**: Valid JSON; schema-valid; all fields 
     match (or only negligible paraphrasing).
   * **4 - Good**: Valid JSON; schema-valid; most fields correct 
     with minor omissions; no contradictions.
   * **3 - Fair**: Valid JSON; schema-valid; some fields correct, 
     but notable errors or omissions.
   * **2 - Poor**: Valid JSON but major mismatches, schema invalid, 
     or reasoning contradicts Ground Truth.
   * **1 - Inadequate**: Not valid JSON, or completely inconsistent.

   Note: Schema invalidation caps the score at 2 maximum.

4. **Output Final JSON**
   * "score": integer 1-5.
   * "justification": brief explanation citing specific issues.

Here are the inputs for you to evaluate:

Ground Truth JSON:
[BEGIN GROUND TRUTH]
{ground_truth}
[END GROUND TRUTH]

SUT Output (to be parsed and validated):
[BEGIN SUT OUTPUT]
{sut_output}
[END SUT OUTPUT]

JSON Schema to validate the SUT Output against:
[BEGIN SCHEMA]
{json_schema}
[END SCHEMA]
\end{lstlisting}

\section{Additional Implementation Details}
\label{app:implementation-details}

\subsection{SQL Prompt Auto-gen}
\label{app:sql-prompt-autogen}

HYVE generates SQL tool prompts directly from the request-scoped datastore rather than from hand-written templates. This design keeps the prompt aligned with the \emph{actual} tables created during preprocessing and avoids inconsistent schema descriptions. At a high level, prompt generation has two goals: to expose the live relational structure that is available for querying, and to provide task-appropriate guidance for the next tool-enabled LLM step.

For the primary LLM call, HYVE appends a unified prompt that exposes the two high-level entry points available at that stage: \texttt{GenTemplateAndBackfill}, which enters Mode~2 template backfill, and \texttt{QueryDatastore}, which executes Mode~3 evidence-query reasoning. The prompt includes a schema description derived from the live datastore together with guidance and examples grounded in the tables, columns, and relations available in the current request. However, the two branches are not disclosed symmetrically. For \texttt{QueryDatastore}, the primary-call prompt already provides the full tool specification and the SQL-oriented guidance needed for the model to issue evidence queries immediately. By contrast, \texttt{GenTemplateAndBackfill} is exposed in the primary call only as an entry-point decision, not with its full backfill instructions. This asymmetry follows the gradual-disclosure design described earlier: detailed prompt instructions are revealed only when they are needed.

If the primary call selects \texttt{GenTemplateAndBackfill}, HYVE generates a second prompt specialized to Mode~2. This follow-up prompt introduces the \texttt{BackfillData} tool contract and provides the detailed instructions and auto-generated examples needed to map queried values back into the model-generated template in a schema-consistent way. If the primary call instead selects \texttt{QueryDatastore}, HYVE enters Mode~3. In this branch, no new SQL prompt is generated, because the full \texttt{QueryDatastore} specification and SQL guidance were already provided in the primary-call prompt. Instead, HYVE executes the SQL tool call, appends the returned evidence to the prior conversation context, and issues the bounded follow-up LLM call for final synthesis described in Section~\ref{sec:bounded}.

\begin{algorithm}[t]
\caption{Request-Specific SQL Prompt Generation}
\label{alg:sql-prompt-generation}
\begin{algorithmic}[1]
\STATE \textbf{Input:} Datastore $\mathcal{D}$, base message $m$, prompt stage $\sigma \in \{\text{primary-call}, \text{mode2}, \text{mode3}\}$
\STATE \textbf{Output:} Tool prompt or follow-up context $\pi$
\STATE Extract the live schema and relational structure from $\mathcal{D}$
\STATE Build a request-specific schema description for the current tables and relations
\IF{$\sigma = \text{primary-call}$}
    \STATE Expose \texttt{GenTemplateAndBackfill} as the Mode~2 entry-point choice
    \STATE Specify the full \texttt{QueryDatastore} contract for Mode~3 evidence-query reasoning
    \STATE Add guidance and examples for evidence-oriented SQL generation
\ELSIF{$\sigma = \text{mode2}$}
    \STATE Specify the \texttt{BackfillData} tool contract
    \STATE Add detailed guidance and examples for schema-consistent template backfilling
\ELSE
    \STATE Execute the \texttt{QueryDatastore} SQL tool call
    \STATE Append the returned evidence to the prior message context
    \STATE Prepare the bounded follow-up LLM call for final synthesis
\ENDIF
\STATE Assemble the resulting prompt or follow-up context $\pi$
\STATE \textbf{return} $\pi$
\end{algorithmic}
\end{algorithm}

Because every prompt is synthesized from the same live schema, both the tool instructions and the few-shot examples are
  grounded in the actual tables, columns, and relations available for the current request, rather than in generic placeholders. The prompt is
  therefore both \emph{schema-grounded} and \emph{branch-aware}: the primary-call prompt fully supports Mode~3 evidence-query reasoning while exposing only
  the Mode~2 entry point, the Mode~2 follow-up prompt is disclosed only after template backfill has been selected, and the Mode~3 follow-up stage reuses
  the SQL guidance already provided in the primary call by appending queried evidence for final synthesis. Together, these features help HYVE guide the LLM
  toward richer and more accurate SQL tool calls.

\subsection{Mode~2: \texttt{BackfillData} Execution}
\label{app:postprocessor}
The postprocessor inspects the primary LLM output and, when the request enters Mode~2, executes the deterministic \texttt{BackfillData} path. In this mode, the LLM generates both a partially visible answer template and the SQL-based backfill specification, including the query and its column-to-template mappings. The postprocessor then executes that specification against the datastore and restores the hidden values while preserving the template structure produced by the model. The task-specific prompt is synthesized from the \emph{actual} datastore schema and teaches the model how SQL results should be mapped back into the partially visible template.

If the datastore contains a multi-series table with a \texttt{series\_idx} column and attached series labels, the prompt may include an example such as:
\begin{lstlisting}[basicstyle=\ttfamily\footnotesize]
{
  "tool_name": "BackfillData",
  "query": "SELECT x, y, series_idx 
            FROM data ORDER BY series_idx, _row_id",
  "mappings": [
    {
      "sql_column": "x",
      "template_path": 
          "path.to.list.(series_idx).data.(N).x"
    },
    {
      "sql_column": "y",
      "template_path": 
          "path.to.list.(series_idx).data.(N).y"
    }
  ]
}
\end{lstlisting}
This means: execute the SQL query on the full datastore, group the returned rows by \texttt{series\_idx}, navigate to the corresponding series object in the template, and rebuild its \texttt{data} array from the SQL values. Here \texttt{(series\_idx)} is not a literal field name in the output; it is a placeholder that the executor replaces with concrete group indices such as \texttt{(0)}, \texttt{(1)}, and \texttt{(2)}. The token \texttt{(N)} denotes the list dimension that should be cleared and rebuilt in the order of the SQL result. This positional convention is necessary because, once the SQL query filters or reorders the original rows, the original \texttt{\_row\_id} values no longer determine the correct positions in the partially visible template. Instead, backfilling must rely on the ordered result sequence itself to reconstruct the list faithfully. Because the rebuilt list items are cloned from the visible template items, the final result preserves the LLM-generated nested schema while replacing only the truncated arrays with full-fidelity datastore values.

\begin{algorithm}[t]
\caption{Multi-Series \texttt{BackfillData} Execution}
\label{alg:multiseries-backfill}
\begin{algorithmic}[1]
\STATE \textbf{Input:} Template JSON $\tau$, SQL query $q$, mappings $M$
\STATE \textbf{Output:} Backfilled output $\tau$
\STATE $R \gets \textsc{ExecuteSQL}(q)$
\STATE $g \gets \textsc{DetectGroupColumn}(M, \textsc{Columns}(R))$
\IF{$g = \textsc{Unsupported}$}
    \RETURN $\tau$
\ELSIF{$g \neq \varnothing$}
    \FOR{each distinct group value $u$ in $R[g]$}
        \STATE $R_u \gets \textsc{RowsWhere}(R, g=u)$
        \FOR{each mapping $(c, p) \in M$}
            \STATE $p_u \gets \textsc{Replace}(p, (g), (u))$
            \STATE $p_u \gets \textsc{ValidateAndCorrectPath}(\tau, p_u)$
            \STATE $V \gets R_u[c]$ in SQL order
            \FOR{each expanded path $\hat{p}$ in $\textsc{ExpandNestedN}(\tau, p_u, |V|)$}
                \STATE $\textsc{BackfillPathWithSchema}(\tau, \hat{p}, V)$
            \ENDFOR
        \ENDFOR
    \ENDFOR
\ELSE
    \FOR{each mapping $(c, p) \in M$}
        \STATE $p \gets \textsc{ValidateAndCorrectPath}(\tau, p)$
        \STATE $V \gets R[c]$ in SQL order
        \FOR{each expanded path $\hat{p}$ in $\textsc{ExpandNestedN}(\tau, p, |V|)$}
            \STATE $\textsc{BackfillPathWithSchema}(\tau, \hat{p}, V)$
        \ENDFOR
    \ENDFOR
\ENDIF
\RETURN $\tau$
\end{algorithmic}
\end{algorithm}

Line~4 detects whether the SQL result contains a grouping field, such as \texttt{series\_idx}, that distinguishes multiple repeated substructures. The algorithm then follows three cases: grouped backfill, flat backfill, or a fallback when grouping is ambiguous or unsupported.

\subsection{Mode~3: Bounded LLM Invocations}
\label{sec:bounded}

HYVE provides an explicit architectural bound on model invocations. After the primary LLM call selects a postprocessing branch, Mode~3 performs, by default, at most one additional LLM call: the system executes the \texttt{QueryDatastore} tool call, appends the returned SQL evidence to a slim context, and asks the model for a final grounded synthesis.

Consequently, HYVE uses at most two LLM calls per request by default: one primary call and, only when necessary, one follow-up call. This keeps latency predictable and is easier to operationalize than open-ended agentic loops, which may issue an unbounded number of tool calls.

As an optional optimization, HYVE also allows a small configurable number of additional repair calls for SQL generation. For example, when a generated query fails due to an execution error or incomplete evidence, the system can supply the error message or missing-context hints and request a revised SQL query. This mechanism remains bounded and is used only to improve query quality under controlled retry limits.

\end{document}